\documentclass{article}
\pdfoutput=1

\usepackage[final,nonatbib]{neurips_2019}
\usepackage[autonum]{tchdr}
\usepackage[sort&compress,numbers]{natbib}
\DeclareMathOperator{\cone}{cone}
\DeclareMathOperator{\conv}{conv}
\newdimen{\algindent}
\makeatletter
\newcommand{\LineComment}[2][0]{\Statex \hspace{#1\algindent} \hskip\ALG@thistlm $\triangleright$ #2}
\makeatother

\usepackage[utf8]{inputenc} 
\usepackage[T1]{fontenc}    

\allowdisplaybreaks

\title{Universal Boosting Variational Inference}

\author{%
  Trevor Campbell\\
  Department of Statistics\\
  University of British Columbia\\
  Vancouver, BC V6T 1Z4 \\
  \texttt{trevor@stat.ubc.ca} 
  \And
  Xinglong Li\\
  Department of Statistics\\
  University of British Columbia\\
  Vancouver, BC V6T 1Z4 \\
  \texttt{xinglong.li@stat.ubc.ca} \\
}

\begin{document}

\maketitle

\begin{abstract}
Boosting variational inference (BVI) approximates an intractable probability density by iteratively
building up a mixture of simple component distributions one at a time, using techniques from sparse convex optimization
to provide both computational scalability and approximation error guarantees. But the guarantees have strong conditions that 
do not often hold in practice, resulting in degenerate component optimization problems; and we show that the ad-hoc regularization 
used to prevent degeneracy in practice can cause BVI to fail in unintuitive ways.
 We thus develop \emph{universal boosting variational inference} (UBVI), a BVI scheme
that exploits the simple geometry of probability densities under the Hellinger metric to prevent
the degeneracy of other gradient-based BVI methods, avoid difficult joint optimizations 
of both component and weight, and simplify fully-corrective weight optimizations. 
 We show that for any target density and any mixture component family, the output of 
UBVI converges to the best possible approximation in the mixture family, even when the mixture family is misspecified.
We develop a scalable implementation based on exponential family mixture components and standard stochastic optimization techniques.
Finally, we discuss statistical benefits of the Hellinger distance as a variational objective
through bounds on posterior probability, moment, and importance sampling errors. Experiments 
on multiple datasets and models show that UBVI provides reliable, accurate posterior approximations.
\end{abstract}

\section{Introduction}
Bayesian statistical models provide a powerful framework for learning from data, with the ability 
to encode complex hierarchical dependence structures and prior domain expertise, as well as coherently
capture uncertainty in latent parameters. The two predominant methods for Bayesian inference are Markov chain Monte Carlo (MCMC) \citep{Brooks11,Gelman13}---which obtains 
approximate posterior samples by simulating a Markov chain---and variational inference (VI) \citep{Jordan99,Wainwright08}---which obtains 
an approximate distribution by minimizing some divergence to the posterior within a tractable family.
The key strengths of MCMC are its generality and the ability to perform a computation-quality tradeoff: one can obtain a higher quality approximation
by simulating the chain for a longer period \citep[Theorem 4 \& Fact 5]{Roberts04}. However, the resulting
Monte Carlo estimators have an unknown bias or random computation time \citep{Jacob17}, and 
statistical distances between the discrete sample posterior approximation and a diffuse true posterior are vacuous, ill-defined, or 
hard to bound without restrictive assumptions or a choice of kernel \citep{Gorham15,Liu16,Chwialkowski16}.
Designing correct MCMC schemes in the large-scale data setting is also a challenging task \citep{Bardenet15,Scott16,Betancourt15}.
VI, on the other hand, is both computationally scalable and widely applicable due to advances from stochastic optimization 
and automatic differentiation \citep{Hoffman13,Kucukelbir17,Ranganath14,Kingma14,Baydin18}. However, the major disadvantage of the approach---and
the fundamental reason that MCMC remains the preferred method in statistics---is that the variational family typically does not
contain the posterior, fundamentally limiting the achievable approximation quality.
And despite recent results in the asymptotic theory of variational methods \citep{Alquier18,Wang18,Yang18,CheriefAbdellatif18,Dehaene18}, it is difficult to assess the effect of the chosen family 
on the approximation for finite data; a poor choice can result in severe underestimation of posterior uncertainty \citep[Ch.~21]{Murphy12}.


Boosting variational inference (BVI) \citep{Guo16,Wang16,Miller17} is an exciting new approach that addresses this fundamental limitation by using a nonparametric mixture variational family.
By adding and reweighting only a single mixture component at a time,
the approximation may be iteratively refined, achieving the computation/quality tradeoff of MCMC and the scalability of VI.
Theoretical guarantees on the convergence rate of Kullback-Leibler (KL) divergence \citep{Guo16,Locatello18,Locatello18b}
are much stronger than those available for standard Monte Carlo, which degrade as the number of estimands increases,
enabling the practitioner to confidently reuse the same approximation for multiple tasks. 
However, the bounds require the KL divergence to be sufficiently smooth over the class of mixtures---an assumption that does not hold
for many standard mixture families, e.g.~Gaussians, resulting in a degenerate procedure in practice. 
To overcome this, an ad-hoc entropy regularization is typically added to each component
optimization; but this regularization invalidates convergence guarantees, and---depending on the regularization weight---sometimes
does not actually prevent degeneracy.

In this paper, we develop \emph{universal boosting variational inference} (UBVI), a variational scheme based on the 
Hellinger distance rather than the KL divergence. The primary advantage of using the Hellinger distance is that it
endows the space of probability densities with a particularly simple unit-spherical geometry in a Hilbert space. 
We exploit this geometry to prevent the degeneracy of other gradient-based BVI methods, avoid difficult joint optimizations 
of both component and weight, simplify fully-corrective weight optimizations, 
and provide a procedure in which the normalization constant of $f$ does not need to be known, 
a crucial property in most VI settings. It also leads to the universality of UBVI: we show that
for \emph{any} target density and \emph{any} mixture component family, the output of UBVI converges to the best possible
approximation in the mixture family, even when the mixture family is misspecified.
We develop a scalable implementation based on exponential family mixture components and standard stochastic optimization techniques.
Finally, we discuss other statistical benefits of the Hellinger distance as a variational objective through bounds 
on posterior probability, moment, and importance sampling errors. 
Experiments 
on multiple datasets and models show that UBVI provides reliable, accurate posterior approximations.




\section{Background: variational inference and boosting}
\begin{figure}[t!]
\begin{center}
\begin{subfigure}[t]{0.47\columnwidth}
\centering\includegraphics[width=0.6\textwidth]{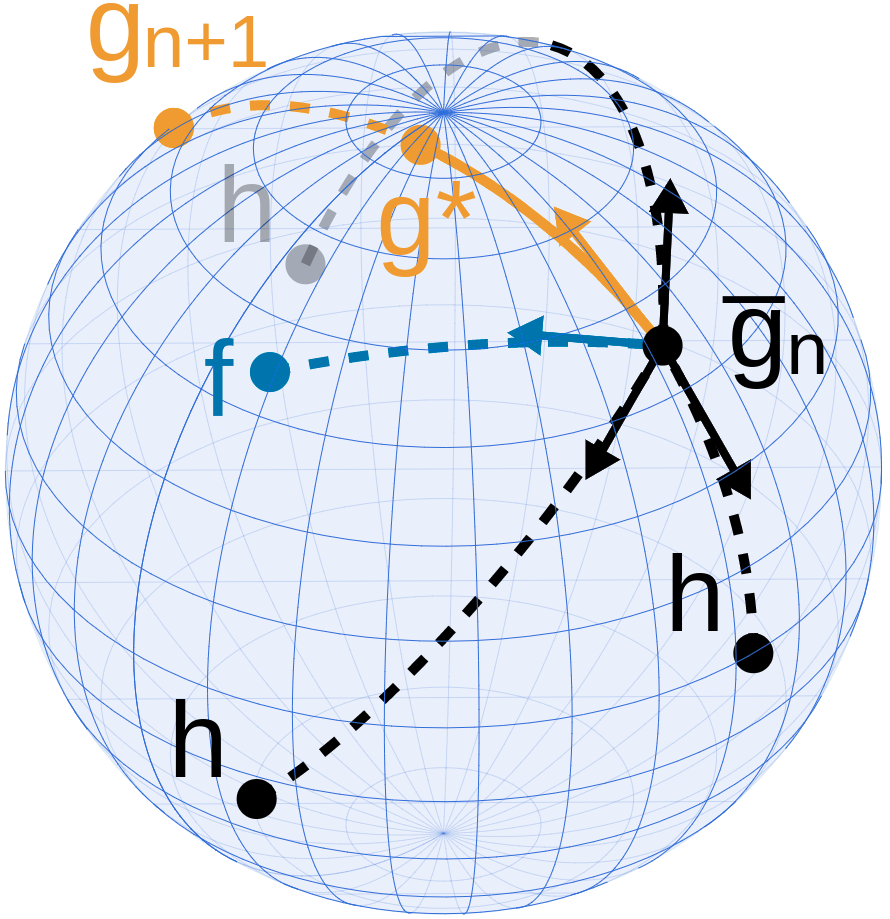}
\caption{}\label{fig:greedy}
\end{subfigure}
\begin{subfigure}[t]{0.47\columnwidth}
\centering\includegraphics[width=0.6\textwidth]{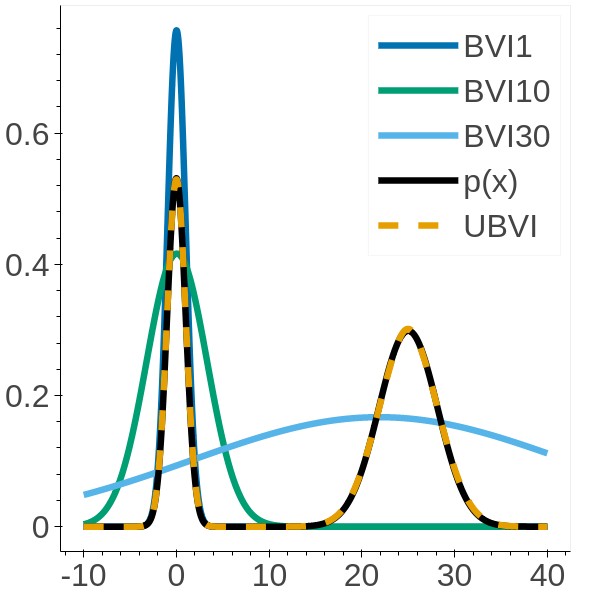}
\caption{}\label{fig:boostfail}
\end{subfigure}
\end{center}
\caption{(\ref{fig:greedy}): Greedy component selection, with target $f$, current iterate $\bg_n$, candidate components $h$, optimal component $g_{n+1}$,
the closest point $g^\star$ to $f$ on the $\bg_n \to g_{n+1}$ geodesic, 
and
arrows for initial geodesic directions. The quality of $g_{n+1}$ is determined by the distance
from $f$ to $g^\star$, or equivalently, by the alignment of the initial directions $\bg_n\to g_{n+1}$ and $\bg_n \to f$.
(\ref{fig:boostfail}): BVI can fail even when $p$ is in the mixture family.
Here $p = \frac{1}{2}\distNorm(0, 1)+\frac{1}{2}\distNorm(25,5)$, and UBVI finds the correct mixture in 2 iterations. 
BVI (with regularization weight in $\{1, 10, 30\}$) does not converge. For example, when the regularization weight is 1 the first component will have variance $< 5$,
and the second component optimization diverges since the target $\distNorm(25, 5)$ component has a heavier tail.
Upon reweighting the second component is removed, and the approximation will never improve.
}
\label{fig:greedyandboostfail}
\end{figure}

Variational inference, in its most general form, involves approximating a
probability density $p$ by minimizing some divergence $\divergence{}{\cdot}{\cdot}$ from $\xi$ to $p$
over densities $\xi$ in a family $\mcQ$,
\[
q = \argmin_{\xi \in \mcQ} \divergence{}{\xi}{p}.
\]
Past work has almost exclusively involved parametric families $\mcQ$, such as mean-field exponential families \citep{Wainwright08}, finite mixtures \citep{Jaakkola98,Zobay14,Gershman12},
normalizing flows \citep{Rezende15}, and neural nets \citep{Kingma14}. The issue with these families
is that typically $\min_{\xi\in\mcQ}\divergence{}{\xi}{p} > 0$---meaning the practitioner cannot achieve arbitrary
approximation quality with more computational effort---and a priori, there is no way to tell how 
poor the best approximation is. To address this, \emph{boosting variational inference} (BVI) \citep{Guo16,Wang16,Miller17}
proposes the use of the nonparametric family of all finite mixtures of a component density family $\mcC$,
\[
\mcQ = \conv{\mcC} &\defined \left\{\sum_{k=1}^Kw_k\xi_k : K\in\nats, w\in\Delta^{K-1}, \,\,\forall k\in\nats\,\,\xi_k \in \mcC\right\}.
\]
Given a judicious choice of $\mcC$, we have that $\inf_{\xi \in \mcQ}\divergence{}{\xi}{p} = 0$; in other words,
we can approximate any continuous density $p$ with arbitrarily low divergence \citep{Parzen62}.
As optimizing directly over the nonparametric $\mcQ$ is intractable, BVI instead adds one component at a time
to iteratively refine the approximation. There are two general formulations of BVI;
Miller et al.~\citep{Miller17} propose minimizing KL divergence over both the weight and component simultaneously,
\[
\!\!\!q_n &= \sum_{k=1}^nw_{nk}\xi_k & \xi_{n+1}, \omega &= \argmin_{\xi\in\mcC, \rho\in[0,1]} \kl{\rho\xi+(1-\rho)q_n}{p} & w_{n+1} = \left[(1-\omega)w_n\,\,\omega\right]^T\!\!\!\!,
\]
while Guo et al.~and Wang \citep{Guo16,Wang16} argue that optimizing both simultaneously is too difficult,
and use a gradient boosting \citep{Schapire90} formulation instead,
\[
\xi_{n+1} &= \argmin_{\xi\in\mcC} \left<\xi, \left.\grad\kl{\cdot}{p}\right|_{q_n}\right> & w_{n+1} &=\argmin_{\omega = \left[(1-\rho)w_n\,\,\rho\right]^T\!\!\!\!,\,\,\rho\in[0,1]}\kl{\sum_{k=1}^{n+1}\omega_k\xi_k}{p}.
\]
Both algorithms attain $\kl{q_N}{p} = O(1/N)$\footnote{We assume throughout that nonconvex optimization problems can be solved reliably.}--- 
the former by appealing to results from convex functional analysis \citep[Theorem II.1]{Zhang03},
and the latter by viewing BVI as functional Frank-Wolfe optimization \citep{Locatello18,Frank56,Jaggi13}. This requires that 
$\kl{q}{p}$ is strongly smooth or has bounded curvature over $q\in\mcQ$, for which it is sufficient that densities in $\mcQ$ are bounded away from 0, bounded above, and have compact support \citep{Locatello18},
or have a bounded parameter space \citep{Locatello18b}. However, these assumptions do not hold in practice for many simple (and common) cases, e.g., where $\mcC$ is
the class of multivariate normal distributions. Indeed, gradient boosting-based BVI methods all require some ad-hoc entropy regularization in the component
optimizations to avoid degeneracy \citep{Guo16,Wang16,Locatello18b}. In particular, given a sequence of regularization weights $r_n > 0$, BVI solves the following
component optimization \citep{Locatello18b}:
\[
\xi_{n+1} &= \argmin_{\xi\in\mcC} \left<\xi, \log\frac{\xi^{r_{n+1}} q_n}{p}\right>.\label{eq:bvi}
\]
This addition of regularization has an adverse effect on performance in practice as
demonstrated in \cref{fig:boostfail}, and can lead to unintuitive behaviour and nonconvergence---even when $p \in \mcQ$ (\cref{prop:gbvi}) or
when the distributions in $\mcC$ have lighter tails than $p$ (\cref{prop:gbvi2}).

\bnprop\label{prop:gbvi}
Suppose $\mcC$ is the set of univariate Gaussians with mean 0 parametrized by variance, let $p = \distNorm(0, 1)$, and let the initial approximation be $q_1 = \distNorm(0, \tau^2)$.
Then BVI in \cref{eq:bvi} with regularization $r_2 > 0$ returns a degenerate
next component $\xi_2$ if $\tau^2 \leq 1$, and iterates infinitely without improving the approximation if $\tau^2 > 1$ and
$r_2 >\tau^2-1$.
\enprop

\bnprop\label{prop:gbvi2}
Suppose $\mcC$ is the set of univariate Gaussians with mean 0 parametrized by variance, and let $p = \distCauchy(0, 1)$.
Then BVI in \cref{eq:bvi} with regularization $r_1 > 0$ returns a degenerate first component $\xi_1$
if $r_1 \geq 2$.
\enprop

\section{Universal boosting variational inference (UBVI)}

\subsection{Algorithm and convergence guarantee}\label{sec:algorithm}
To design a BVI procedure without the need for ad-hoc regularization, we use a variational objective based on
the \emph{Hellinger distance}, which
for any probability space $(\mcX, \Sigma, \mu)$ and densities $p, q$ is
\[
\hell[2]{p}{q} &\defined \frac{1}{2}\int \left(\sqrt{p(x)} - \sqrt{q(x)}\right)^2\mu(\dee x).
\]
Our general approach relies on two facts about the Hellinger distance. First,
the metric $\hell{\cdot}{\cdot}$ endows the set of $\mu$-densities with a simple geometry
corresponding to the nonnegative functions on the unit sphere in $L^2(\mu)$.
 In particular,
if $f, g \in L^2(\mu)$ satisfy $\|f\|_2=\|g\|_2=1$, $f, g \geq 0$, 
then $p=f^2$ and $q=g^2$ are probability densities and
\[
\hell[2]{p}{q} = \frac{1}{2}\left\|f - g\right\|_2^2. 
\]
One can thus perform Hellinger distance boosting 
by iteratively finding components that minimize geodesic distance to $f$ on the unit sphere in $L^2(\mu)$.
Like the Miller et al.~approach \citep{Miller17}, the boosting step directly minimizes a statistical distance,
leading to a nondegenerate method; but like the Guo et al.~and Wang approach \citep{Guo16,Wang16}, this
avoids the joint optimization of both component and weight; see \cref{sec:geoboosting} for details.
Second, a conic combination $g = \sum_{i=0}^N \lambda_i g_i$, $\lambda_i\geq 0$, $\|g_i\|_2=1$, $g_i\geq 0$ in $L^2(\mu)$ satisfying $\|g\|_2=1$
corresponds to the mixture model density
\[
q = g^2 &= \sum_{i, j=1}^N Z_{ij} \lambda_i \lambda_j \left(\frac{g_i g_j}{Z_{ij}}\right) & Z_{ij} \defined \left<g_i, g_j\right> \geq 0.\label{eq:gqdef}
\]
Therefore, if we can find a conic combination satisfying $\|f - g\|_2 \leq \sqrt{2}\epsilon$ for $p = f^2$, we can guarantee that
the corresponding mixture density $q$ satisfies $\hell{p}{q} \leq \epsilon$.
The mixture will be built from a family $\mcH\subset L^2(\mu)$ of component functions 
for which $\forall h \in \mcH$, $\|h\|_2 = 1$ and $h \geq 0$.
We assume that the target function $f \in L^2(\mu)$, $\|f\|_2 = 1$, $f\geq 0$ is known up to proportionality. 
We also assume that $f$ is not orthogonal to $\spann{\mcH}$ for expositional
brevity, although the algorithms and theoretical results presented here apply equally well in this case. 
We make no other assumptions;  in particular, we do not assume $f$ is in $\closure{\spann{\mcH}}$.

\begin{algorithm}[t!]
\caption{The universal boosting variational inference (UBVI) algorithm.}\label{alg:ubvi}
{\small
\begin{algorithmic}[1]
\Procedure{UBVI}{$p$, $\mcH$, $N$}
\State $f \overset{\propto}{\gets} \sqrt{p}$
\State $\bg_{0} \gets 0$
\For{$n=1, \dots, N$}
\LineComment[1]{Find the next component to add to the approximation using \cref{eq:geogreedy}}
\State $g_{n} \gets \argmax_{h\in\mcH}   \left<f - \left<f,\bg_{n-1}\right>\bg_{n-1},h\right>\scalebox{.75}{\bigg/}\sqrt{1-\left<h,\bg_{n-1}\right>^2}$ 
\LineComment{Compute pairwise normalizations using \cref{eq:gqdef}}
\For{$i=1, \dots, n$}
\State $Z_{n, i} = Z_{i,n} \gets \left<g_{n}, g_{i}\right>$ 
\EndFor
\LineComment{Update weights using  \cref{eq:fullcorrectdual}}
\State $d = \left( \left<f, g_1\right>, \dots, \left<f, g_{n}\right> \right)^T$ 
\State $\beta = \argmin_{b \in \reals^{n}, b\geq 0} b^TZ^{-1}b + 2b^TZ^{-1}d$ 
\State $(\lambda_{n,1}, \dots, \lambda_{n,n}) = \frac{Z^{-1}(\beta + d)}{\sqrt{(\beta+d)^TZ^{-1}(\beta+d)}}$ 
\LineComment{Update boosting approximation}
\State $\bg_{n} \gets \sum_{i=1}^{n} \lambda_{ni}g_i$
\EndFor
\State \Return $q = \bg^2_N$
\EndProcedure
\end{algorithmic}
}
\end{algorithm}

The universal boosting variational inference (UBVI) procedure is shown in \cref{alg:ubvi}. In each iteration,
the algorithm finds a new mixture component from $\mcH$ (line 5; see \cref{sec:geoboosting} and \cref{fig:greedy}). 
Once the new component is found, the algorithm solves a convex quadratic optimization problem to update the weights (lines 9--11).
The primary requirement to run \cref{alg:ubvi} is the ability to compute or estimate $\left<h, f\right>$ and $\left<h, h'\right>$ for $h, h'\in\mcH$. 
For this purpose we employ an exponential component family $\mcH$ such that $Z_{ij}$ is available in closed-form, and use samples from $h^2$
to obtain estimates of $\left<h, f\right>$; see \cref{sec:optimization} for further implementation details.

The major benefit of UBVI is that it comes with a computation/quality tradeoff akin to MCMC:
for any target $p$ and component family $\mcH$, (1) there is a unique mixture $\hp=\hf^2$ minimizing
$\hell{\hp}{p}$ over the closure of finite mixtures $\closure{\mcQ}$;
and (2) the output $q$ of UBVI$(p, \mcH, N)$ satisfies $\hell{q}{\hp} = O(1/N)$ with a dimension-independent constant.
No matter how coarse the family $\mcH$ is, the output of UBVI will converge to the best possible mixture approximation.
\cref{thm:convergence} provides the precise result.
\bnthm\label{thm:convergence}
For any density $p$ there is a unique density $\hp=\hf^2$ satisfying $\hp = \argmin_{\xi\in\closure{\mcQ}}\hell{\xi}{p}$.
If each component optimization \cref{eq:geogreedy} is solved with a relative suboptimality of at most $(1-\delta)$, then
the variational mixture approximation $q$ returned by \emph{UBVI}($p$, $\mcH$, $N$) satisfies
\[
\hell{\hp}{q}^2 &\leq \frac{J_1}{1+\left(\frac{1-\delta}{\tau}\right)^2 J_1 (N-1)} & J_1 &\defined 1-\left<\hf, g_1\right>^2\in[0,1) & \tau&\defined\text{\cref{eq:tau}} < \infty.
\]
\enthm
The proof of \cref{thm:convergence} may be found in \cref{sec:convergenceproofs}, and consists of three primary steps. 
First, \cref{lem:hf} guarantees the existence and uniqueness of the convergence target $\hf$ under possible misspecification of the component family $\mcH$.
Then the difficulty of approximating $\hf$ with conic combinations of functions in $\mcH$ is 
captured by the basis pursuit denoising problem \citep{Chen01}
\[
\tau \defined \inf_{\begin{subarray}{c} h_i \in \cone\mcH\\ x\in[0,1)\end{subarray}} \quad &(1-x)^{-1}\sum_{i=1}^\infty \|h_i\|_2 \quad
\text{s.t.}\quad  \|\hf - \sum_{i=1}^\infty h_i\|_2\leq x, \quad \forall i,\, h_i\geq 0.\label{eq:tau}
\]
\cref{lem:tau} guarantees that $\tau$ is finite, and in particular $\tau \leq \frac{\sqrt{1-J_1}}{1-\sqrt{J_1}}$, which can be estimated in practice
using \cref{eq:hellestimators}. Finally, \cref{lem:descent} develops an objective function recursion, which is then solved to yield \cref{thm:convergence}. 
Although UBVI and \cref{thm:convergence} is reminiscent of past work on greedy approximation in a 
Hilbert space \citep{Schapire90,Freund97,Barron08,Chen10,Mallat93,Chen89,Chen99,Tropp04,Campbell18b},
it provides the crucial advantage that the greedy steps do not require knowledge of the normalization of $p$.
UBVI is inspired by a previous greedy method \citep{Campbell18b}, but provides guarantees with
an arbitrary, potentially misspecified infinite dictionary in a Hilbert space, and uses quadratic optimization to perform weight updates.
Note that both the theoretical and practical cost of UBVI is dominated by finding the next component (line 5), 
which is a nonconvex optimization problem.
The other expensive step is inverting $Z$; however, incremental methods using block matrix inversion \citep[p.~46]{Petersen12}
reduce the cost at iteration $n$ to $O(n^2)$ and overall cost to $O(N^3)$, which is not a concern for practical mixtures with $\ll 10^3$ components. 
The weight optimization (line 10) is a nonnegative least squares problem, which can be solved efficiently \citep[Ch.~23]{Lawson95}.

\subsection{Greedy boosting along density manifold geodesics}\label{sec:geoboosting}
This section provides the technical derivation of UBVI (\cref{alg:ubvi}) by expoiting the geometry of 
square-root densities under the Hellinger  metric.
Let the conic combination in $L^2(\mu)$ after initialization followed by $N-1$ steps of greedy construction be denoted 
\[
\bg_{n} \defined \sum_{i=1}^n \lambda_{ni} g_i, \quad \|\bg_n\|_2 = 1,
\]
where $\lambda_{ni} \geq 0$ is the weight for component $i$ at step $n$, and $g_i$ is the component added at step $i$.
%
%
To find the next component, we minimize the distance between $\bg_{n+1}$ and $f$ over choices of $h\in\mcH$ and position $x\in[0,1]$ along the $\bg_n \to h$ geodesic,\footnote{Note that the $\argmax$ may not be unique, and when $\mcH$ is infinite it may not exist; \cref{thm:convergence} still holds and UBVI works as intended in this case.
For simplicity, we use $(\dots) = \argmax (\dots)$ throughout.}
\[
\bg_0 = 0 \qquad g_{n+1}, x^\star &= \argmin_{h\in\mcH, x\in[0,1]} \left\|f - \left(x\frac{h - \left<h, \bg_n\right>\bg_n}{\|h-\left<h, \bg_n\right>\bg_n\|_2}+\sqrt{1-x^2}\bg_n\right)\right\|_2\label{eq:protogreedy}\\
&= \argmax_{h\in\mcH, x\in[0,1]} x \left<f, \frac{h - \left<h, \bg_n\right>\bg_n}{\|h-\left<h, \bg_n\right>\bg_n\|_2}\right> +\sqrt{1-x^2}\left<f,\bg_n\right>.
\]
Noting that $h-\left<h,\bg_n\right>\bg_n$ is orthogonal to $\bg_n$, the second term does not depend on $h$, and $x\geq 0$,
we avoid optimizing the weight and component simultaneously and find that
\[
g_{n+1} &= \argmax_{h\in\mcH} \left<\frac{f - \left<f, \bg_n\right>\bg_n}{\|f-\left<f, \bg_n\right>\bg_n\|_2}, \frac{h - \left<h, \bg_n\right>\bg_n}{\|h-\left<h, \bg_n\right>\bg_n\|_2}\right>= \argmax_{h\in\mcH} \frac{\left<f - \left<f, \bg_n\right>\bg_n, h\right>}{\sqrt{1-\left<h, \bg_n\right>^2}}.\label{eq:geogreedy}
\]
 Intuitively, \cref{eq:geogreedy} attempts to maximize alignment of $g_{n+1}$ with the residual $f - \left<f,\bg_n\right>\bg_n$ (the numerator)
resulting in a ring of possible solutions, and among these, \cref{eq:geogreedy} minimizes alignment with the current iterate $\bg_n$ (the denominator).
The first form in \cref{eq:geogreedy} provides an alternative intuition:
$g_{n+1}$ achieves the maximal alignment of the \emph{initial geodesic directions} $\bg_n \to f$ and $\bg_n \to h$ on the sphere. 
See \cref{fig:greedy} for a depiction.
After selecting the next component $g_{n+1}$,
one option to obtain $\bg_{n+1}$ is to use the optimal weighting $x^\star$ from \cref{eq:protogreedy};
in practice, however, it is typically the case that solving \cref{eq:geogreedy} is expensive enough that finding
the optimal set of coefficients for $\{g_1, \dots, g_{n+1}\}$ is worthwhile. This is accomplished by maximizing
alignment with $f$ subject to a nonnegativity and unit-norm constraint: 
\[
(\lambda_{(n+1)1}, \dots, \lambda_{(n+1)(n+1)}) = \argmax_{x\in\reals^{n+1}}\quad & \left< f, \sum_{i=1}^{n+1} x_i g_i\right> \quad \text{s.t.} \quad  x\geq 0, \quad x^TZx \leq 1,
\label{eq:fullcorrect}
\]
where $Z\in\reals^{N+1\times N+1}$ is the matrix with entries $Z_{ij}$ from \cref{eq:gqdef}.
Since projection onto the feasible set of \cref{eq:fullcorrect} may be difficult, the problem
may instead be solved using the dual via
\[
\begin{aligned}
&(\lambda_{(n+1)1}, \dots, \lambda_{(n+1)(n+1)}) = \frac{Z^{-1}(\beta + d)}{\sqrt{(\beta+d)^TZ^{-1}(\beta+d)}}\\
&d = \left( \left<f, g_1\right>, \dots, \left<f, g_{n+1}\right> \right)^T\qquad
\beta = \argmin_{b \in \reals^{n+1}, b\geq 0} b^TZ^{-1}b + 2b^TZ^{-1}d.
\end{aligned}\label{eq:fullcorrectdual}
\]
\cref{eq:fullcorrectdual} is a nonnegative linear least-squares problem---for which very efficient algorithms are available \citep[Ch.~23]{Lawson95}---in 
contrast to prior variational boosting methods, where a fully-corrective weight update is a general constrained convex optimization problem. 
Note that, crucially, none of the above steps rely on knowledge of the normalization constant of $f$.

\section{Hellinger distance as a variational objective}\label{sec:hellinger}

\begin{figure}[t!]
\begin{subfigure}{0.45\textwidth}
    \centering\includegraphics[width=0.75\textwidth]{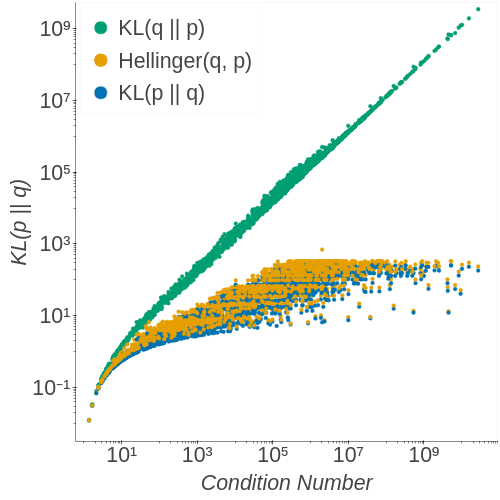}
\caption{}\label{fig:klcompare_kl}
\end{subfigure}
\begin{subfigure}{0.45\textwidth}
    \centering\includegraphics[width=0.75\textwidth]{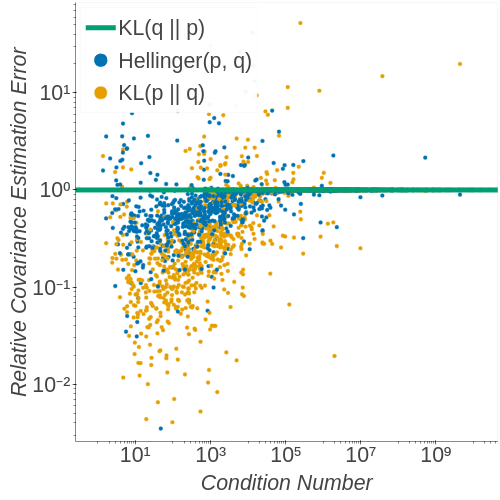}
\caption{}\label{fig:klcompare_coverr}
\end{subfigure}
\caption{Forward KL divergence---which controls worst-case downstream importance sampling error---and importance-sampling-based covariance estimation error on
a task of approximating $\distNorm(0, A^TA)$, $A_{ij}\distiid \mathcal{N}(0, 1)$ with $\mathcal{N}(0, \sigma^2I)$ by minimizing Hellinger, forward KL, and reverse KL,
plotted as a function of condition number $\kappa(A^TA)$. Minimizing Hellinger distance provides significantly lower forward KL divergence and estimation error than minimizing reverse KL.}\label{fig:klcompare}
\end{figure}

While the Hellinger distance has most frequently been applied in asymptotic analyses (e.g., \citep{Ghosal00}), 
it has seen recent use as a variational objective \citep{Li16} and possesses a number of particularly
useful properties that make it a natural fit for this purpose. First, $\hell{\cdot}{\cdot}$ 
 applies to any arbitrary pair of densities, unlike $\kl{p}{q}$, which requires that $p \ll q$.
Minimizing $\hell{\cdot}{\cdot}$ also implicitly minimizes error in 
posterior probabilities and moments---two quantities of primary importance to practitioners---via its control
on total variation and $\ell$-Wasserstein by \cref{lem:tvdbd,prop:hellwass}. Note that
the upper bound in \cref{lem:tvdbd} is typically tighter 
than that provided by the usual $\kl{q}{p}$ variational objective via Pinsker's inequality (and at the very least is always in $[0, 1]$),
and the bound in \cref{prop:hellwass} 
shows that convergence in $\hell{\cdot}{\cdot}$ implies convergence in up to $\ell^\text{th}$ moments \citep[Theorem 6.9]{Villani} under 
relatively weak conditions.
\bnprop[{e.g.~\citep[p.~61]{Pollard}}]\label{lem:tvdbd}
The Hellinger distance bounds total variation via
\[
\hell[2]{p}{q} &\leq \tvd{p}{q} \defined \frac{1}{2}\|p-q\|_1 \leq \hell{p}{q}\sqrt{2 -\hell[2]{p}{q}}.\label{eq:htvbd}
\]
\enprop
\bnprop\label{prop:hellwass}
Suppose $\mcX$ is a Polish space with metric $d(\cdot, \cdot)$, $\ell \geq 1$, 
and $p, q$ are densities with respect to a common measure $\mu$. Then for any $x_0$,
\[
W_\ell(p, q)
&\leq 2\hell{p}{q}^{\nicefrac{1}{\ell}} \left(\EE\left[d(x_0, X)^{2\ell}\right]+\EE\left[d(x_0, Y)^{2\ell}\right]\right)^{\nicefrac{1}{2\ell}},
\]
where $Y\dist p(y)\mu(\dee y)$ and $X\dist q(x)\mu(\dee x)$. In particular, if densities $(q_N)_{N\in\nats}$ and $p$
have uniformly bounded $2\ell^\text{th}$ moments, $\hell{p}{q_N}\to 0 \implies W_\ell(p, q_N) \to 0$
as $N\to\infty$.
\enprop
Once a variational approximation $q$ is obtained, it will typically be used to estimate
expectations of some function of interest $\phi(x)\in L^2(\mu)$ via Monte Carlo. 
Unless $q$ is trusted entirely, this
involves importance sampling---using $I_n(\phi)$ or $J_n(\phi)$ in \cref{eq:imptcsamplingests} depending on whether 
the normalization of $p$ is known---to account for the error in $q$ compared with the target distribution $p$  \citep{Yao18},
\[
I_n(\phi) &\defined \frac{1}{N}\sum_{n=1}^N \frac{p(X_i)}{q(X_i)}\phi(X_i) & J_n(\phi) &\defined \frac{I_n(\phi)}{I_n(1)} & X_i&\distiid q(x)\mu(\dee x).\label{eq:imptcsamplingests}
\]
Recent work has shown that the error of importance sampling is controlled by
the intractable forward KL-divergence $\kl{p}{q}$ \citep{Chatterjee18}. This is where the Hellinger distance shines;
\cref{prop:hellkl} shows that it penalizes both positive and negative values of $\log p(x)/q(x)$ and thus 
provides moderate control on $\kl{p}{q}$---unlike $\kl{q}{p}$, which only penalizes negative values.  See \cref{fig:klcompare} for a demonstration
of this effect on the classical correlated Gaussian example \citep[Ch.~21]{Murphy12}. While the takeaway from this setup is typically that minimizing $\kl{q}{p}$ may cause severe 
underestimation of variance, a reasonable practitioner should attempt to use importance sampling to correct for this anyway. But \cref{fig:klcompare}
shows that minimizing $\kl{q}{p}$ doesn't minimize $\kl{p}{q}$ well, leading to poor estimates from importance sampling.
Even though  minimizing $\hell{p}{q}$ also underestimates variance, it provides enough control on $\kl{p}{q}$ so that importance sampling
can correct the errors. Direct bounds on the error of importance sampling estimates are also provided in \cref{prop:hellimptc}.
\bnprop\label{prop:hellkl}
Define $R \defined \log \frac{p(X)}{q(X)}$ where $X \dist p(x)\mu(\dee x)$. Then
\[
\hell{p}{q}&
\geq \frac{1}{2}\sqrt{\EE\left[R^2\left(\frac{1+\ind\left[R\leq 0\right]R}{1+R}\right)^2\right]} \geq \frac{\kl{p}{q}}{2\sqrt{1 + \EE\left[\ind\left[R>0\right](1+R)^2\right]}}.
\]
\enprop
\bnprop\label{prop:hellimptc}
Define $\alpha \defined \left(N^{-\nicefrac{1}{4}} + 2\sqrt{\hell{p}{q}}\right)^2$.
Then the importance sampling error with known normalization is bounded by
\[
\EE\left[\left|I_n(\phi) - I(\phi)\right|\right] &\leq \|\sqrt{p}\phi\|_{2}\alpha,
\]
and with unknown normalization by
\[
\forall t > 0 \qquad \Pr( |J_n(\phi)-I(\phi)| > \|\sqrt{p}\phi\|_{2}t) &\leq \left(1+4t^{-1}\sqrt{1+t}\right)\alpha.
\]
\enprop
Next, the Hellinger distance between densities $q, p$ can be estimated with high relative accuracy given samples from $q$,
enabling the use of the above bounds in practice. This involves computing either $\widehat{\hell[2]{p}{q}}$ or $\widetilde{\hell[2]{p}{q}}$ below,
depending on whether the normalization of $p$ is known.
The expected error of both of these estimates relative to $\hell{p}{q}$ is bounded via \cref{prop:hellest}.
\[
	\hspace{-.3cm}\widehat{\hell[2]{p}{q}} &\defined 1\!-\!\frac{1}{N}\sum_{n=1}^N \sqrt{\frac{p(X_n)}{q(X_n)}}, & \!\!\widetilde{\hell[2]{p}{q}} &\defined 1\!-\! \frac{\frac{1}{N}\sum_{n=1}^N \sqrt{\frac{p(X_n)}{q(X_n)}}}{\sqrt{\frac{1}{N}\sum_{n=1}^N \frac{p(X_n)}{q(X_n)}}}, & \!\!X_n &\distiid q(x)\mu(\dee x). \label{eq:hellestimators}
\]
\bnprop\label{prop:hellest}
The mean absolute difference between the Hellinger squared estimates is
\[
\EE\left[ \left| \widehat{\hell[2]{p}{q}} - \hell{p}{q}^2\right| \right] &\leq 
 \frac{\hell{p}{q}\sqrt{2-\hell{p}{q}^2}}{\sqrt{N}}\\
\EE\left[ \left| \widetilde{\hell[2]{p}{q}}- \hell{p}{q}^2\right| \right] &\leq 
\sqrt{2}\left(1+\sqrt{N}^{-1}\right)\hell{p}{q}.
\]
\enprop

It is worth pointing out that although the above statistical properties of the Hellinger distance make it well-suited
as a variational objective, it does pose computational issues during optimization. In particular, to avoid 
numerically unstable gradient estimation, one must transform Hellinger-based objectives such as \cref{eq:geogreedy}.
This typically produces biased and occasionally higher-variance Monte Carlo gradient estimates than the corresponding KL 
gradient estimates. We detail these transformations and other computational considerations in \cref{sec:optimization}.

\section{Experiments}\label{sec:expts}
\renewcommand{\UrlFont}{\ttfamily}
In this section, we compare UBVI, KL divergence boosting variational inference (BVI) \cite{Locatello18b}, 
BVI with an ad-hoc stabilization in which $q_n$ in \cref{eq:bvi} is replaced by $q_n + 10^{-3}$ to help prevent degeneracy (BVI+),
 and standard VI. For all experiments, we used a regularization schedule of $r_n = 1/\sqrt{n}$ for BVI(+) in \cref{eq:bvi}.
We used the multivariate Gaussian family for $\mcH$ parametrized by mean and log-transformed diagonal covariance matrix.
We used 10,000 iterations of ADAM \citep{Kingma15}  for optimization, with decaying step size $1/\sqrt{1+i}$
and Monte Carlo gradients based on 1,000 samples.
Fully-corrective weight optimization was conducted via simplex-projected SGD for BVI(+) and nonnegative least squares for UBVI.
Monte Carlo estimates of $\left<f, g_n\right>$ in UBVI were based on 10,000 samples.
Each component optimization was initialized from the best of 10,000 trials of sampling a component (with mean $m$ and covariance $\Sigma$) from the current mixture, sampling the initialized component mean from $\distNorm(m, 16\Sigma)$,
and setting the initialized component covariance to $\exp(Z)\Sigma$, $Z\dist\distNorm(0, 1)$.
Each experiment was run 20 times with an Intel i7 8700K processor and 32GB of memory.
Code is available at  \url{www.github.com/trevorcampbell/ubvi}.

\begin{figure}[t!]
\begin{center}
\begin{subfigure}[t]{0.3\columnwidth}
\centering\includegraphics[width=\textwidth]{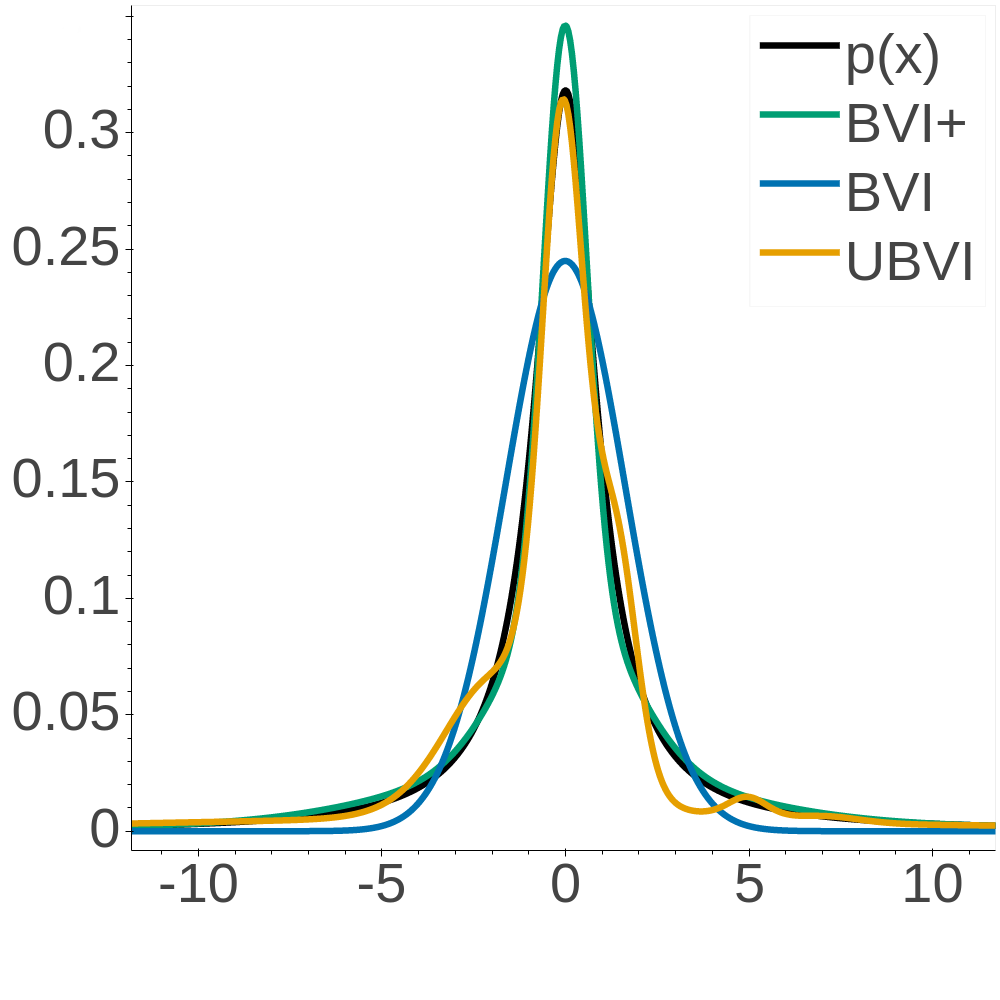}
\caption{Cauchy density}\label{fig:cauchy}
\end{subfigure}
\begin{subfigure}[t]{0.3\columnwidth}
\centering\includegraphics[width=\textwidth]{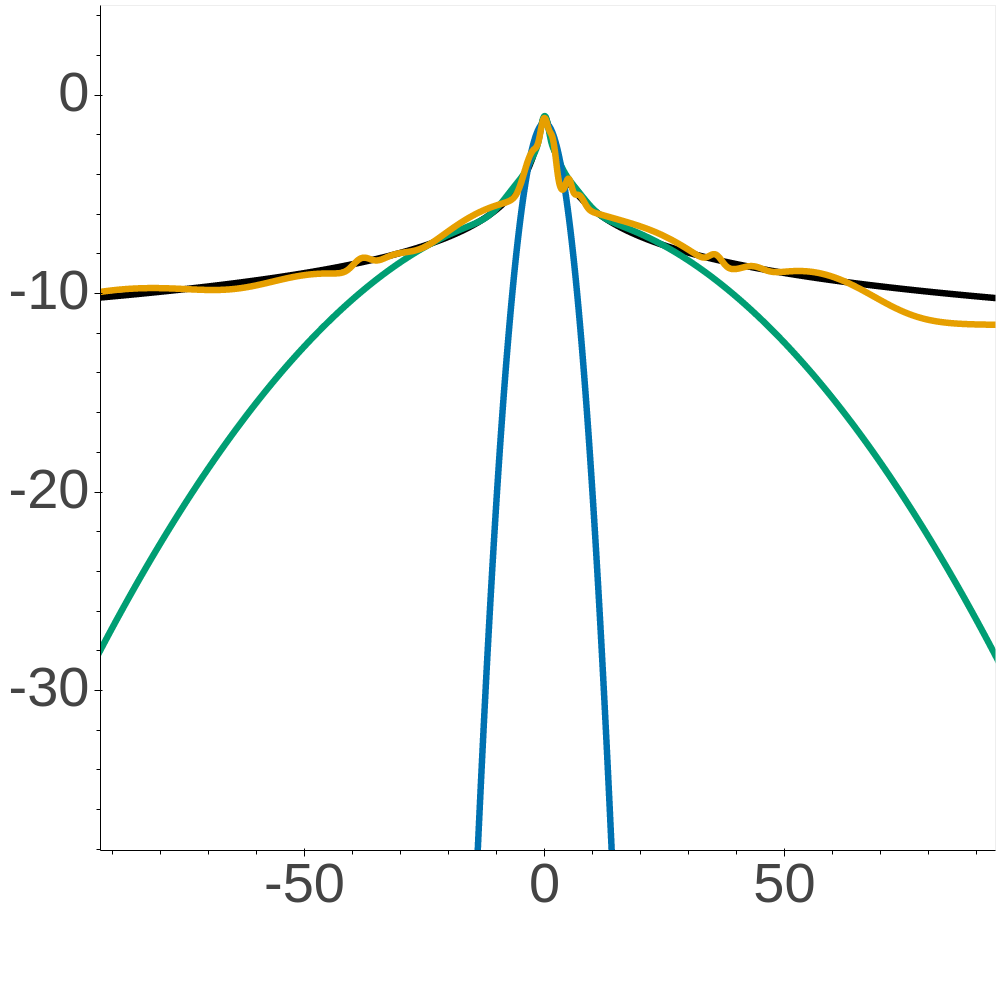}
\caption{Cauchy log density}\label{fig:log_cauchy}
\end{subfigure}
\begin{subfigure}[t]{0.3\columnwidth}
\centering\includegraphics[width=\textwidth]{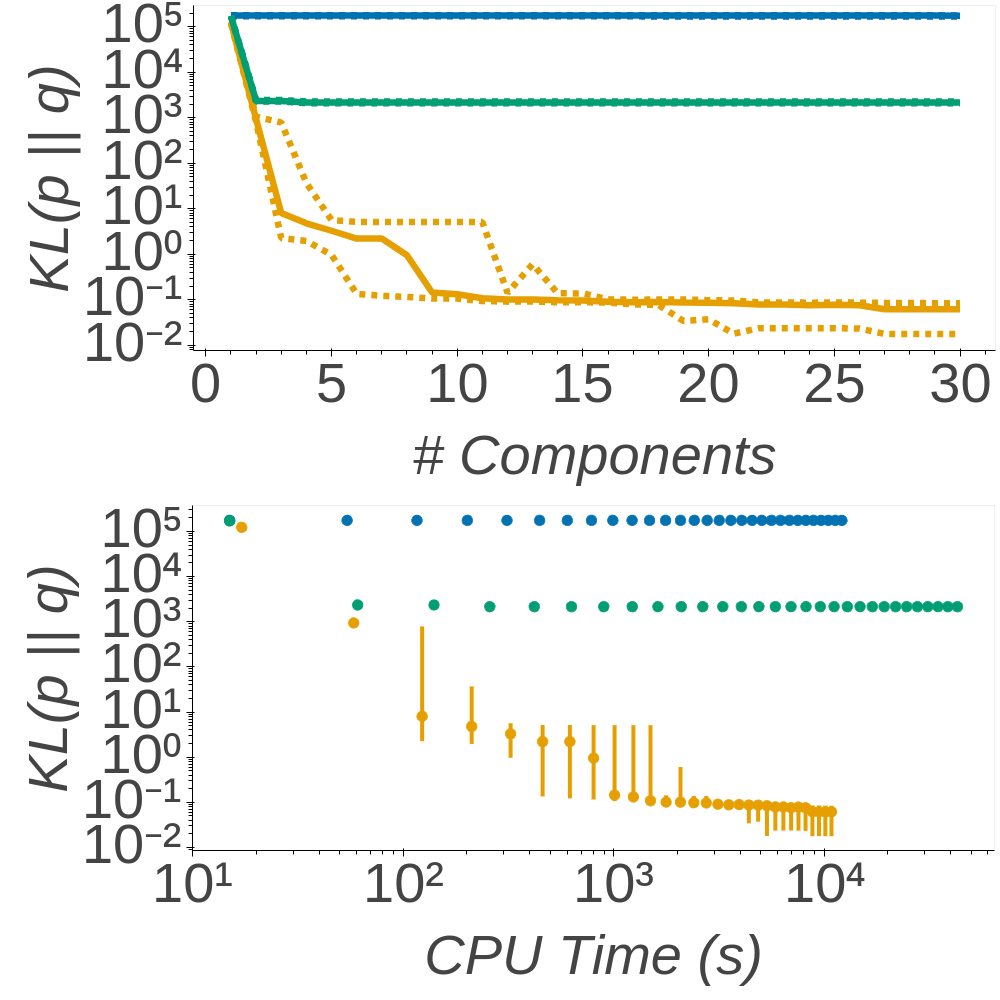}
\caption{Cauchy KL divergence}\label{fig:cauchy_kl}
\end{subfigure}

\begin{subfigure}[t]{0.3\columnwidth}
\centering\raisebox{0.7cm}{\includegraphics[width=\textwidth]{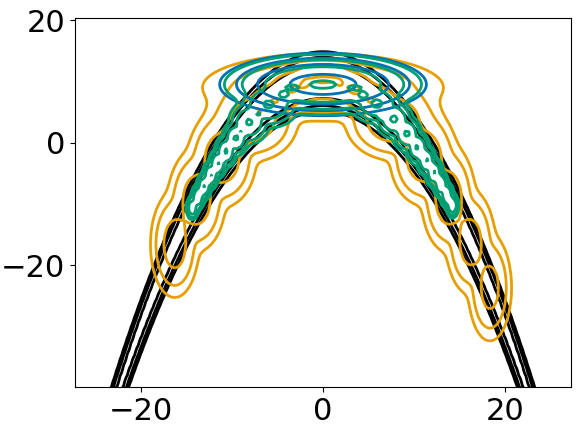}}
\caption{Banana density}\label{fig:banana}
\end{subfigure}
\begin{subfigure}[t]{0.3\columnwidth}
\centering\includegraphics[width=\textwidth]{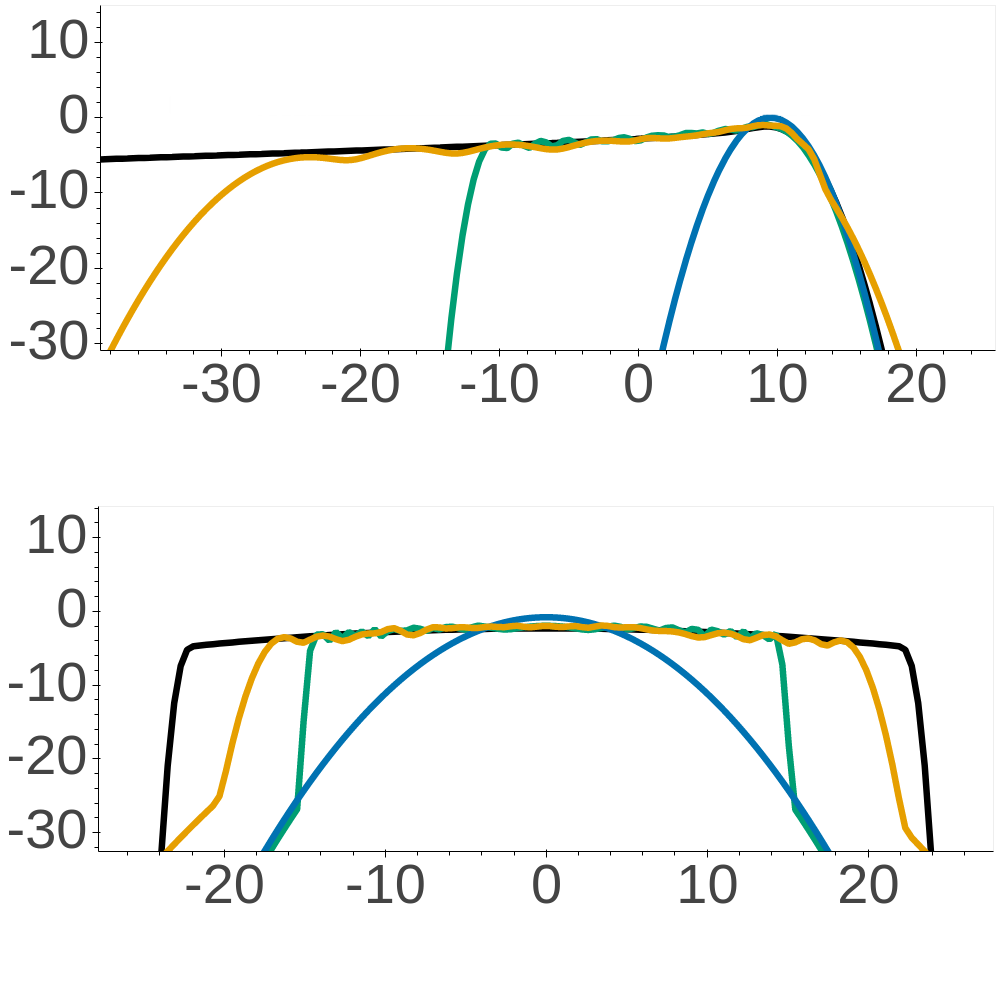}
\caption{Banana log marginal densities}\label{fig:log_banana}
\end{subfigure}
\begin{subfigure}[t]{0.3\columnwidth}
\centering\includegraphics[width=\textwidth]{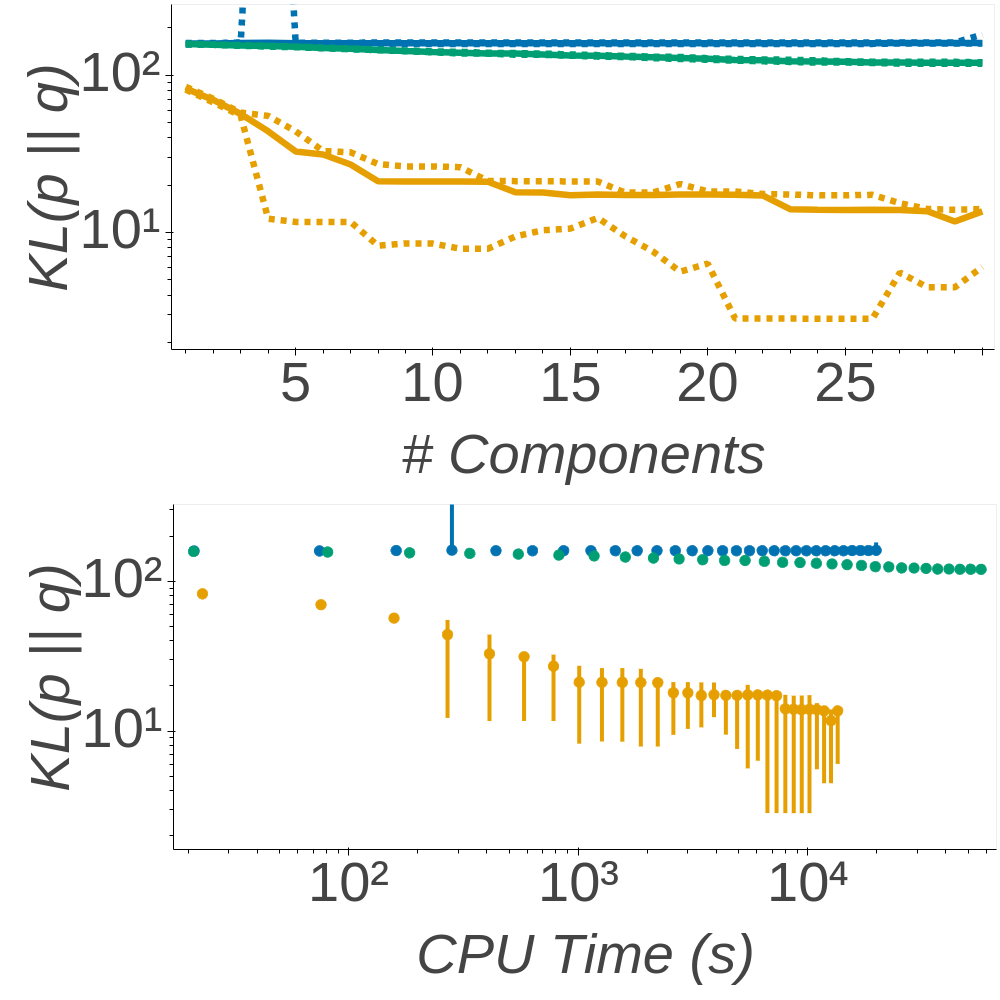}
\caption{Banana KL divergence}\label{fig:banana_kl}
\end{subfigure}

\end{center}
\caption{Results on the Cauchy and banana distributions; all subfigures use the legend from \cref{fig:cauchy}.
(\cref{fig:cauchy,fig:banana}): Density approximation with 30 components for Cauchy (\ref{fig:cauchy}) and banana (\ref{fig:banana}).
BVI has degenerate component optimizations after the first, while UBVI and BVI+ are able to refine the approximation.
(\cref{fig:log_cauchy,fig:log_banana}): Log density approximations for Cauchy (\ref{fig:log_cauchy})
and banana marginals (\ref{fig:log_banana}). UBVI provides more accurate approximation
of distribution tails than the KL-based BVI(+) algorithms.
(\cref{fig:cauchy_kl,fig:banana_kl}): The forward KL divergence 
 vs.~the number of boosting components and computation time. UBVI consistently improves its approximation
as more components are added, while the KL-based BVI(+) methods improve either slowly or not at all due to degeneracy.
Solid lines / dots indicate median, and dashed lines / whiskers indicate $25^\text{th}$ and $75^\text{th}$ percentile.
}
\label{fig:banana_cauchy_experiments}
\end{figure}

\subsection{Cauchy and banana distributions}\label{sec:cauchy_banana_expt}
\cref{fig:banana_cauchy_experiments} shows the results of running UBVI, BVI, and BVI+ for 30 boosting iterations
 on the standard univariate Cauchy distribution 
and the banana distribution \citep{Haario01} with curvature $b=0.1$.
These distributions were selected for their heavy tails and complex structure (shown in \cref{fig:log_cauchy,fig:log_banana}), two
features that standard variational inference does not often address but boosting methods
should handle. However, BVI particularly struggles with heavy-tailed distributions, where its 
component optimization objective after the first is degenerate. BVI+ is able to refine its approximation,
but still cannot capture heavy tails well, leading to large forward KL divergence (which controls downstream
importance sampling error). We also found that the behaviour of BVI(+) is very sensitive
to the choice of regularization tuning schedule $r_n$, and is difficult to tune well.
 UBVI, in contrast, approximates both heavy-tailed
and complex distributions well with few components, and involves no tuning effort beyond the component optimization 
step size. 

%
%

\begin{figure}[t!]
\begin{center}
\begin{subfigure}[t]{0.3\columnwidth}
\centering\includegraphics[width=\textwidth]{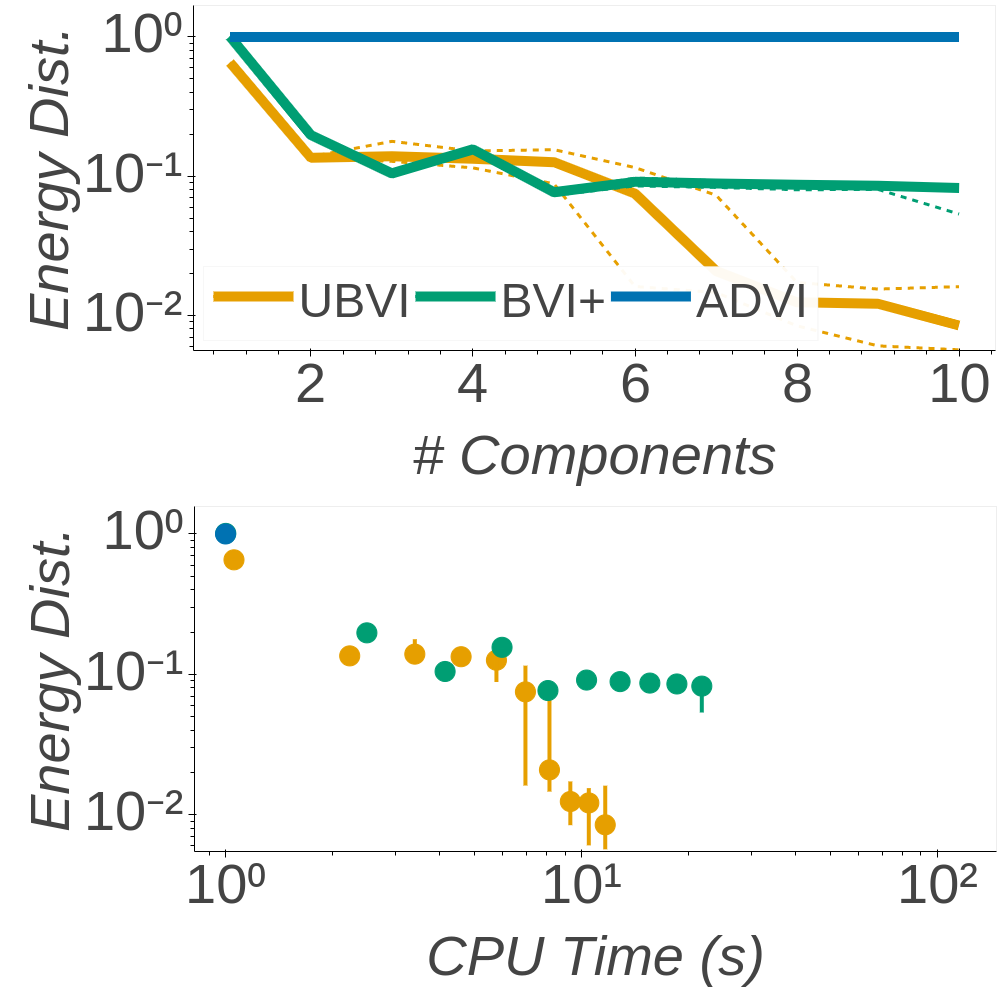}
\caption{Synthetic}\label{fig:synth}
\end{subfigure}
\begin{subfigure}[t]{0.3\columnwidth}
\centering\includegraphics[width=\textwidth]{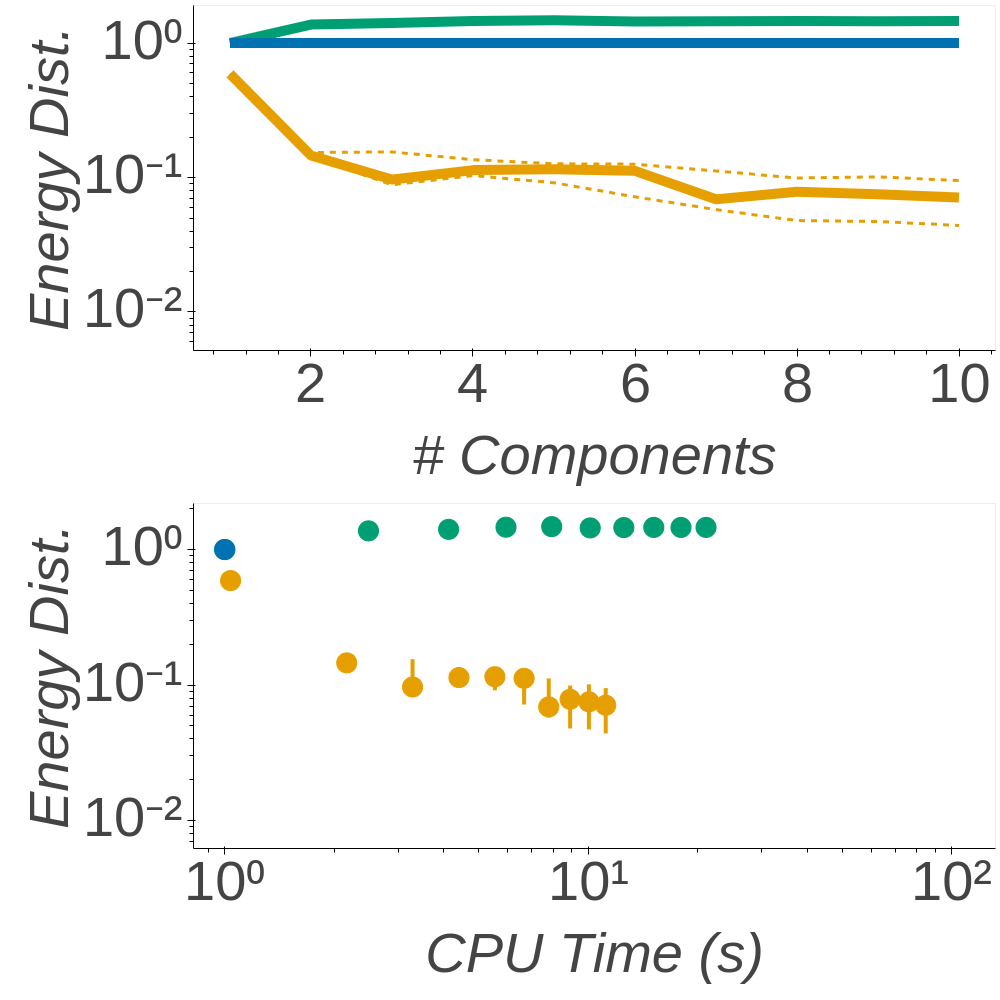}
\caption{Chemical Reactivity}\label{fig:ds1}
\end{subfigure}
\begin{subfigure}[t]{0.3\columnwidth}
\centering\includegraphics[width=\textwidth]{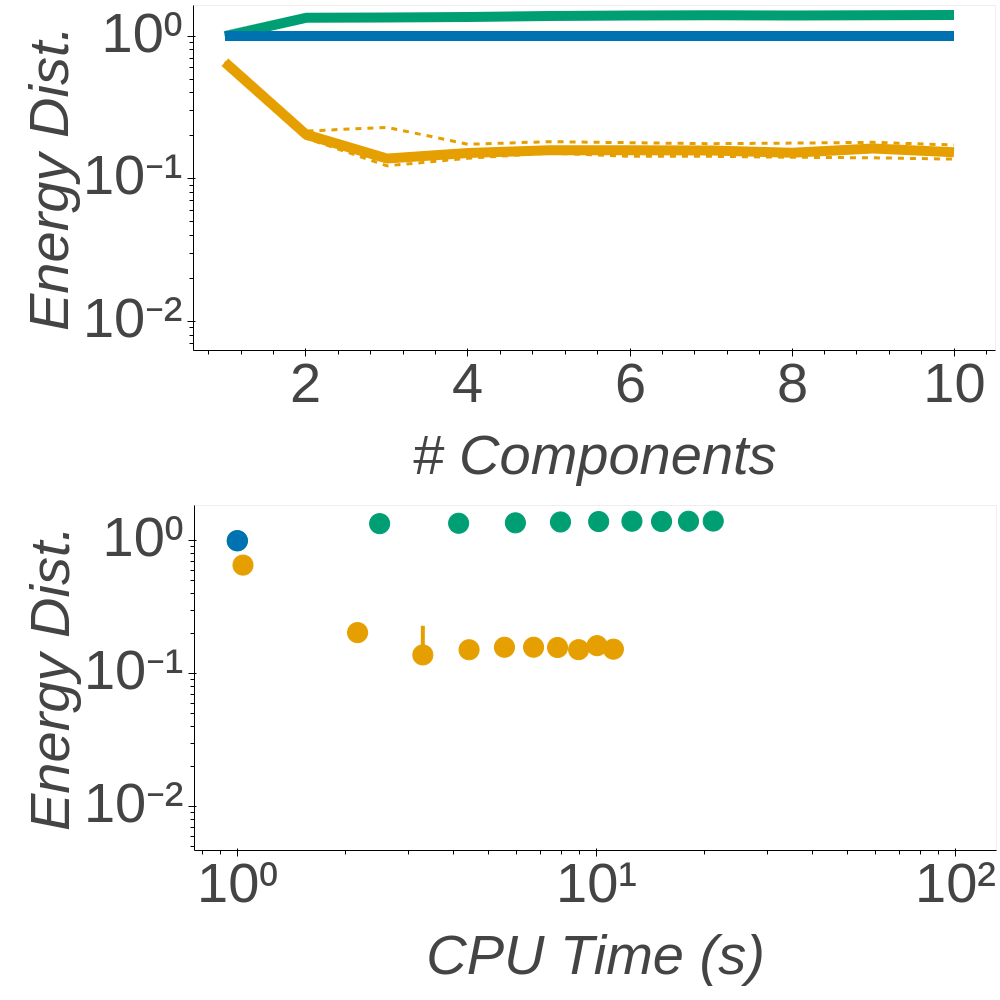}
\caption{Phishing}\label{fig:phish}
\end{subfigure}
\end{center}
\caption{Results from Bayesian logistic regression posterior inference on the synthetic (\ref{fig:synth}), chemical (\ref{fig:ds1}), and phishing (\ref{fig:phish}) datasets, showing the 
energy distance \citep{Szekely13} to the posterior (via NUTS \citep{Hoffman14}) vs.~the number of 
components and CPU time. Energy distance and time are normalized by the VI median.
Solid lines / dots indicate median, and dashed lines / whiskers indicate $25^\text{th}$ / $75^\text{th}$ percentile.
}
\label{fig:logistic_experiment}
\end{figure}

\subsection{Logistic regression with a heavy-tailed prior}
\cref{fig:logistic_experiment} shows the results of running 10 boosting iterations of  UBVI, BVI+, and standard VI
for posterior inference in Bayesian logistic regression. We used a multivariate $\distT_2(\mu, \Sigma)$ prior, 
where in each trial, the prior parameters were set via $\mu = 0$ and $\Sigma = A^TA$ for $A_{ij}\distiid \distNorm(0, 1)$.
 We ran this experiment on a 2-dimensional synthetic dataset generated from the model, a 10-dimensional chemical reactivity dataset,
and a 10-dimensional phishing websites dataset, each with 20 subsampled datapoints.\footnote{Real datasets available online at  \url{http://komarix.org/ac/ds/} and \url{https://www.csie.ntu.edu.tw/~cjlin/libsvmtools/datasets/binary.html}.} The small dataset size and heavy-tailed prior were chosen to create a complex posterior structure better-suited to evaluating boosting variational methods. The results in \cref{fig:logistic_experiment} are similar to those in the synthetic test from \cref{sec:cauchy_banana_expt}; UBVI is able to refine its posterior approximation as it adds components without tuning effort, while the KL-based BVI+ method is difficult to tune well and does not reliably provide better posterior approximations than standard VI. BVI (no stabilization) is not shown, as its component optimizations after the first are degenerate and it reduces to standard VI.

\section{Conclusion}
This paper developed universal boosting variational inference (UBVI). UBVI optimizes the Hellinger metric,
avoiding the degeneracy, tuning, and difficult joint component/weight optimizations of other gradient-based BVI methods,
while simplifying fully-corrective weight optimizations. Theoretical guarantees on the convergence of Hellinger distance
provide an MCMC-like computation/quality tradeoff, and experimental results demonstrate the advantages
over previous variational methods.

\section{Acknowledgments}
T.~Campbell and X.~Li are supported by a National Sciences and Engineering Research Council of Canada (NSERC) 
Discovery Grant and an NSERC Discovery Launch Supplement.

\small
\bibliographystyle{unsrt}
\bibliography{sources}

\clearpage
\appendix
\section{Component family and practical aspects of optimization}\label{sec:optimization}
In order to use \cref{alg:ubvi}, we need to compute or estimate
$\left<h, \phi\right>$ for any $\phi\in L^2(\mu)$ and $\left<h, h'\right>$ for any $h, h'\in\mcH$. 
For arbitrary $\phi$, we use Monte Carlo estimates based on samples from $h^2$ via
\[
\forall\phi\in L^2(\mu), \qquad \left<h, \phi\right> &= \int h^2(x) \frac{\phi(x)}{h(x)}\mu(\dee x) = \EE_{h^2}\left[\frac{\phi(X)}{h(X)}\right] \approx \frac{1}{S}\sum_{s=1}^S \frac{\phi(X_s)}{h(X_s)}\quad X_s\distiid h^2,
\]
and employ an exponential component family $\mcH$ such that inner products $\left<h, h'\right>$ between members of $\mcH$ are available in closed-form.
In other words, for some base density $k(x)$, sufficient statistic $T(x)$, and log-partition $A(\eta)$, we let
\[
\mcH = \left\{h_\eta \in L^2(\mu) : h_\eta^2(x) = k(x)\exp\left(\eta^T T(x) - A(\eta)\right)\right\}.
\]
Denoting $\eta_i$ to be the natural parameter for $g_i$, then $g_i = h_{\eta_i}$ and
\[
Z_{ij}=\left<g_i, g_j\right> &= \int h_{\eta_i}(x)h_{\eta_j}(x) \mu(\dee x)
= \exp\left( A\left(\frac{\eta_i+\eta_j}{2}\right)-\frac{A(\eta_i) + A(\eta_j)}{2}\right).\label{eq:zijeval}
\]

In practice, we use a few techniques to improve the stability and performance of UBVI:
\paragraph{Component Initialization} The performance of variational boosting methods is often sensitive to the choice 
of initialization in each component optimization. The initialization used in this work is based on the intuition
that after the first component optimization, each subsequent optimization will typically do one of two things: either it will
find a new mode, or it will attempt to refine a previously found mode. If we wish to refine a previous mode, it is useful
to initialize the optimization near that mode with a similar covariance structure. If we wish to discover a new mode, it
is preferable to sample an initialization from the present distribution with significant added noise. In the experimental
section of this work, we take the middle ground. We first sample a component from the current mixture approximation.
Then, we generate an initialization for the Gaussian mean by sampling from that component with its covariance increased by a factor of 16.
Finally, we initialize the covariance by using that component's covariance multiplied by a standard log-normal random variable.
\paragraph{Objective Transformation} We maximize $\log(J(x))\ind\left[J(x)\geq 0\right] - \log(-J(x))\ind\left[J(x) < 0\right]$, where $J(x)$ is the objective in \cref{eq:geogreedy},
to avoid vanishing gradients and handle possible negativity; while this technically
makes the Monte Carlo-based stochastic gradient estimates biased, it significantly improves performance in practice.
\paragraph{Parametrization} The choice of parametrization can have a significant effect on the conditioning of the optimization problem.
Although we exploit the properties of the exponential family for $Z_{ij}$ evaluation in \cref{eq:zijeval}, we do not use
the natural parametrization during optimization.
In particular, we optimize over the mean and log-transformed marginal variances $\log \sigma_i^2$ in the diagonal covariance matrix $\Sigma = \diag(\sigma_1^2, \dots, \sigma_D^2)$.
\paragraph{Large-Scale Data} If the target density $p$ arises from a Bayesian posterior inference problem with a large dataset, computing
$p$ and its gradients exactly in each component optimization iteration is expensive. Thus, one can use a Monte Carlo minibatch approximation
with uniformly subsampled data per \citep{Li16}. 
\paragraph{Estimating $\left<f, g\right>$} We use different numbers of samples for the component optimization stochastic gradient estimates and the estimates of
$\left<f, g_n\right>$ (Line 9, \cref{alg:ubvi}) required to solve the UBVI weight optimization. In particular, we use a relatively
high number of samples (10,000 in our experiments) for estimating $\left<f, g_n\right>$, as these each need to be estimated only once, and they have
a high impact on the choice of weights and thus future components; and for stochastic optimization, we use a lower number of samples (1,000 in our experiments)
to avoid overly expensive component optimizations.

\section{Proofs}\label{sec:proofs}
\subsection{Proof of gradient boosting BVI behaviour}\label{sec:gbviproofs}

\bprfof{\cref{prop:gbvi}}
Let $\phi(x; \sigma^2)$ be the normal density with mean $0$ and variance $\sigma^2$. Then 
\cref{eq:bvi} is
\[
\sigma^{\star2}=&\argmin_{\sigma^2} \int \phi(x; \sigma^2)\log\frac{\phi(x; \sigma^2)^{r_2} \phi(x; \tau^2)}{\phi(x; 1)} \dee x\\
=& \argmin_{\sigma^2} -r_2 \log \sigma - \frac{\sigma^2}{2\tau^2} + \frac{\sigma^2}{2}\\
=& \left\{ \begin{array}{ll}
\infty & \tau^2 \leq 1\\
 \frac{r_2\tau^2}{ \tau^2-1}& \tau^2 > 1.
\end{array}\right.
\]
Therefore, if the initialization has variance $\tau^2 \leq 1$ the component optimization is degenerate. 
Note that for any two variances $\sigma_1^2$, $\sigma_2^2$, the weight optimization is
\[
w^\star &= \argmin_{w \in [0, 1]} \int \left( w \phi(x; \sigma_1^2) + (1-w) \phi(x; \sigma_2^2)\right) \log \frac{\left( w \phi(x; \sigma_1^2) + (1-w) \phi(x; \sigma_2^2)\right)}{\phi(x; 1)}\dee x\\
&=\argmin_{w \in [0, 1]} w\frac{\sigma_1^2}{2} - w \frac{\sigma_2^2}{2} +  \int \left( w \phi(x; \sigma_1^2) + (1-w) \phi(x; \sigma_2^2)\right) \log \left( w \phi(x; \sigma_1^2) + (1-w) \phi(x; \sigma_2^2)\right)\dee x,
\]
and taking first and second derivatives,
\[
\der{}{w} &= \frac{\sigma_1^2 - \sigma_2^2}{2} + \int \left( \phi(x; \sigma_1^2) - \phi(x; \sigma_2^2)\right) \log \left( w \phi(x; \sigma_1^2) + (1-w) \phi(x; \sigma_2^2)\right)\dee x\\
\der{^2}{w^2}&= \int\frac{\big(\phi(x; \sigma_1^2) - \phi(x;\sigma_2^2)\big)^2}{w\phi(x;\sigma_1^2) + (1-w)\phi(x;\sigma_2^2)}\dee x > 0
\]
At $w=1$, $\int \left( \phi(x; \sigma_1^2) - \phi(x; \sigma_2^2)\right) \log \left( w \phi(x; \sigma_1^2) + (1-w) \phi(x; \sigma_2^2)\right)\dee x = (\sigma_2^2 - \sigma_1^2)/(2\sigma_1^2)$. Therefore, if $\sigma_2^2 > \sigma_1^2 > 1$, 
\[
\der{}{w} < \frac{\sigma_1^2 - \sigma_2^2}{2} + \frac{\sigma_2^2 - \sigma_1^2}{2\sigma_1^2} <0.
\]
In other words, the derivative is always negative, so the optimization sets $w=1$ and forgets the new component. This situation occurs 
if $\sigma_1^2 = \tau^2 >1$, $\sigma_2^2 = r_2\frac{\tau^2}{\tau^2-1}$ and 
$r_2 > \tau^2 - 1$.
\eprfof

\bprfof{\cref{prop:gbvi2}}
Using the notation from the proof of \cref{prop:gbvi}, \cref{eq:bvi} is
\[
\sigma^{\star2}=&\argmin_{\sigma^2} \int \phi(x; \sigma^2)\log\frac{\phi(x; \sigma^2)^{r_1} }{\distCauchy(x; 0, 1)} \dee x\\
=& \argmin_{\sigma^2} -\frac{1}{2}r_1\log\sigma^2 + \EE_{\distNorm(0, 1)}\left[\log(1+\sigma^2x^2)\right].
\]
Taking the derivative with respect to $\sigma^2$ followed by Jensen's inequality yields
\[
	\der{}{\sigma^2} &= \sigma^{-2}\left(-\frac{1}{2}r_1 + \EE_{\distNorm(0, 1)}\left[\frac{\sigma^2x^2}{1+\sigma^2x^2}\right]\right)\\ 
	&\leq \sigma^{-2}\left(-\frac{1}{2}r_1 + \frac{\sigma^2}{1+\sigma^2}\right).
\]
Therefore if $r_1 \geq 2$, the derivative with respect to $\sigma^2$ is always negative, so $\sigma^2$ increases without bound. 
\eprfof

\subsection{Proofs of Hellinger distance properties}\label{sec:hellingerproofs}
\bprfof{\cref{lem:tvdbd}}
This follows from
\[
\hell[2]{p}{q} = \frac{1}{2}\int(f(x)-g(x))^2\mu(\dee x) &\leq \frac{1}{2}\int\left|f(x)-g(x)\right|(f(x)+g(x))\mu(\dee x)\\
&= \frac{1}{2}\int\left|f^2(x)-g^2(x)\right|\mu(\dee x) = \tvd{p}{q}
\]
and
\[
\tvd{p}{q} &= \frac{1}{2}\int\left|f^2(x)-g^2(x)\right|\mu(\dee x) \\
&= \frac{1}{2}\int |f(x)-g(x)|(f(x)+g(x))\mu(\dee x)\\
&\leq \frac{1}{2}\sqrt{\int |f(x)-g(x)|^2\mu(\dee x)\int (f(x)+g(x))^2\mu(\dee x)}\\
 &= \frac{1}{\sqrt{2}}\hell{p}{q}\sqrt{2 + 2\int f(x)g(x)\mu(\dee x)}\\
 &= \hell{p}{q}\sqrt{2 -\hell[2]{p}{q}}.
\]
\eprfof

\bprfof{\cref{prop:hellwass}}
Combining a bound on the $\ell$-Wasserstein distance \citep[Theorem 6.15]{Villani}, 
\[
W^\ell_\ell(p, q) &\leq 2^{\ell-1} \int d(x_0, x)^\ell \left| p(x) - q(x)\right| \mu(\dee x),
\]
with $|p(x)-q(x)| = |\sqrt{p(x)}-\sqrt{q(x)}|(\sqrt{p(x)}+\sqrt{q(x)})$,
Cauchy-Schwarz, and \cref{lem:tvdbd} implies
\[
W^\ell_\ell(p, q)
&\leq 2^{\ell-1/2}\hell{p}{q}\sqrt{\int d(x_0, x)^{2\ell} \left(\sqrt{p(x)}+\sqrt{q(x)}\right)^2 \mu(\dee x)}.
\]
Finally, since $(a+b)^2 \leq 2a^2+2b^2$ for $a, b \in \reals$, 
\[
W^\ell_\ell(p, q)
&\leq 2^{\ell}\hell{p}{q}\sqrt{\int d(x_0, x)^{2\ell}(p(x)+q(x))\mu(\dee x)}.
\]
\eprfof
\bprfof{\cref{prop:hellkl}}
Rearranging the definition of Hellinger distance squared,
\[
\hell[2]{p}{q} &= \frac{1}{2}\int \left(\sqrt{p(x)}-\sqrt{q(x)}\right)^2\mu(\dee x)\\
&= \frac{1}{2}\int p(x) \frac{q(x)}{p(x)}\left(\sqrt{\frac{p(x)}{q(x)}}-1\right)^2\mu(\dee x).
\]
For $x > 1$, $x^{-1}\left(\sqrt{x} - 1\right)^2 \geq \left(\frac{\log x}{1+\log x}\right)^2$, 
and for $x \leq 1$, $x^{-1}\left(\sqrt{x} - 1\right)^2 \geq \left(\log x\right)^2$, so
\[
\hell[2]{p}{q} &\geq 
\frac{1}{2}\int_{p > q} p(x)\left(\frac{\log \frac{p(x)}{q(x)}}{1+ \log \frac{p(x)}{q(x)}} \right)^2\mu(\dee x) +\frac{1}{2}\int_{p \leq q} p(x) \left(\log\frac{p(x)}{q(x)}\right)^2\mu(\dee x).
\]
Now using the relation $2a^2 + 2b^2 \geq (a+b)^2$, 
\[
\hell[2]{p}{q} &\geq 
\frac{1}{4}\int p(x)\left(\ind\left[p>q\right]\frac{\log \frac{p(x)}{q(x)}}{1+ \log \frac{p(x)}{q(x)}} + \ind\left[p\leq q\right]\log\frac{p(x)}{q(x)}\right)^2\mu(\dee x)\\
&= \frac{1}{4}\int p(x)\left(\frac{\ind\left[p>q\right]\log \frac{p(x)}{q(x)}+ \ind\left[p\leq q\right]\log\frac{p(x)}{q(x)}\left(1+ \log \frac{p(x)}{q(x)}\right)}{1+ \log \frac{p(x)}{q(x)}} \right)^2\mu(\dee x)\\
&= \frac{1}{4}\int p(x)\left(\log\frac{p(x)}{q(x)}\right)^2\left(\frac{1+ \ind\left[\log \frac{p(x)}{q(x)}\leq 0\right]\log\frac{p(x)}{q(x)}}{1+ \log \frac{p(x)}{q(x)}} \right)^2\mu(\dee x).
\]
This provides the first result.
Using the reverse H\"older inequality $\|f g\|_1 \geq \|f\|_{\frac{1}{p}}\|g\|_{\frac{-1}{p-1}}$ for $p = 2 \in (1, \infty)$,
\[
\hell[2]{p}{q} &\geq \frac{1}{4}\left(\int p(x) \log\frac{p(x)}{q(x)}\mu(\dee x)\right)^{2}\left(\int p(x)\left(\frac{1+ \log \frac{p(x)}{q(x)}}{1+ \ind\left[\log \frac{p(x)}{q(x)}\leq 0\right]\log\frac{p(x)}{q(x)}}\right)^{2} \mu(\dee x) \right)^{-1}\\
 &= \frac{1}{4}\kl[2]{p}{q}\left(\Pr\left(\log \frac{p(x)}{q(x)} \leq 0\right) + \int p(x)\ind\left[\log \frac{p(x)}{q(x)} > 0\right]\left(1+ \log \frac{p(x)}{q(x)}\right)^{2} \mu(\dee x) \right)^{-1}\\
 &\geq \frac{1}{4}\kl[2]{p}{q}\left(1 + \int p(x)\ind\left[\log \frac{p(x)}{q(x)} > 0\right]\left(1+ \log \frac{p(x)}{q(x)}\right)^{2} \mu(\dee x) \right)^{-1}.
\]
\eprfof
\bprfof{\cref{prop:hellimptc}}
This proof uses a technique adapted from \citep[Theorem 1.1]{Chatterjee18}.
Let $Y \dist p(y)\mu(\dee y)$, $X\dist q(x)\mu(\dee x)$, and for $a \geq 0$,
\[
\rho(x) &\defined \left|1-\sqrt{\frac{q(x)}{p(x)}}\right|^2 & h(x) &\defined \phi(x)\ind\left[\rho(x) \leq a\right].
\]
Then by  Cauchy-Schwarz, 
\[
\EE\left[\left|I_n(\phi) - I_n(h)\right|\right] &\leq \|\phi\|_{L^2(p)}\sqrt{\Pr\left(\rho(Y) > a\right)} \label{eq:bd1}\\
\left|I(\phi) - I(h)\right| &\leq \|\phi\|_{L^2(p)}\sqrt{\Pr\left(\rho(Y) > a\right)}\label{eq:bd2}\\
\EE\left[\left|I_n(h) - I(h)\right|\right] &\leq \sqrt{N}^{-1} \sqrt{\var{\frac{p(X)}{q(X)} \phi(X)\ind\left[\rho(X) \leq a\right] }}.\label{eq:bd3}
\]
Now note that
\[
\var{\left(\frac{p(X)}{q(X)} \phi(X)\ind\left[\rho(X) \leq a\right]\right)} &\leq \EE\left[\frac{p^2(X)}{q^2(X)} \phi(X)^2\ind\left[\rho(X) \leq a\right]\right]\\
&= \EE\left[\frac{p(Y)}{q(Y)} \phi(Y)^2 \ind\left[\rho(Y) \leq a\right]\right]
\]
and for $a \in [0, 1)$, $\rho(x) \leq a$ implies
\[
\sqrt{\frac{q(x)}{p(x)}} &\geq 1-\sqrt{a} \implies
\frac{p(x)}{q(x)} \leq (1-\sqrt{a})^{-2}
\]
So
\[
\var{\left(\frac{p(X)}{q(X)} \phi(X)\ind\left[\rho(X) \leq a\right]\right)} &\leq \|\phi\|^2_{L^2(p)} \left(\frac{1}{(1-\sqrt{a})^2}\right)
\]
and hence
\[
\EE\left[\left|I_n(h) - I(h)\right|\right] &\leq \|\phi\|_{L^2(p)}\sqrt{N}^{-1}\frac{1}{1-\sqrt{a}}
\]
By Markov's inequality,
\[
\Pr\left(\rho(Y) > a\right) &\leq a^{-1}\EE\left[\rho(Y)\right] = 2a^{-1}\hell[2]{p}{q}.
\]
So substituting and combining the three bounds from \cref{eq:bd1,eq:bd2,eq:bd3} using the triangle inequality,
\[
\EE\left[\left|I_n(\phi) - I(\phi)\right|\right] &\leq \|\phi\|_{L^2(p)}\left( \frac{1}{\sqrt{N}(1-\sqrt{a})} + \sqrt{8a^{-1}}\hell{p}{q}\right).
\]
Optimizing over $a$ yields
\[
\sqrt{a} = \frac{8^{1/4}\hell{p}{q}^{1/2}}{8^{1/4}\hell{p}{q}^{1/2} + N^{-1/4}},
\]
and substituting with $8^{1/4} \leq 2$ yields
\[
\EE\left[\left|I_n(\phi) - I(\phi)\right|\right] &\leq \|\phi\|_{L^2(p)}\left(N^{-\nicefrac{1}{4}} + 2\sqrt{\hell{p}{q}}\right)^2.
\]
Setting $N = \alpha^{-4} \hell{p}{q}^{-2}$ yields the first result.
For the second, note that
$|I_n(\phi) - I(\phi)| \leq \|\phi\|_{L^2(p)}\delta$ and $|I_n(1) - 1| \leq \eta$ implies that
\[
\left|J_n(\phi) - I(\phi)\right| = \frac{\left|I_n(\phi) - I_n(1)I(\phi)\right|}{I_n(1)} &\leq\frac{\left|I_n(\phi) - I(\phi)\right| + I(\phi)\left|I_n(1) - 1\right|}{1 - \left|I_n(1) - 1\right|}\\
\leq\|\phi\|_{L^2(p)}\frac{\delta + \eta}{1 - \eta},
\]
so
\[
\Pr\left(\left|J_n(\phi) - I(\phi)\right| > \|\phi\|_{L^2(p)}\frac{\delta + \eta}{1 - \eta}\right) &\leq
\Pr\left(\left|I_n(\phi) - I(\phi)\right| > \|\phi\|_{L^2(p)}\delta\right)+
\Pr\left(\left|I_n(\phi) - 1\right| > \eta\right),
\]
which by Markov inequality and the previous bound,
\[
\Pr\left(\left|J_n(\phi) - I(\phi)\right| > \|\phi\|_{L^2(p)}\frac{\delta + \eta}{1 - \eta}\right) &\leq
\left(N^{-\nicefrac{1}{4}} + 2\sqrt{\hell{p}{q}}\right)^2
\left(\delta^{-1} + \eta^{-1}\right)
\]
Minimizing $\delta^{-1}+\eta^{-1}$ subject to the constraint that $t = (\delta+\eta)/(1-\eta)$ yields the result.
%
\eprfof

\bprfof{\cref{prop:hellest}}
For the first bound, by Jensen's inequality
\[
\EE\left[ \left| \widehat{\hell[2]{p}{q}} - \hell[2]{p}{q}\right| \right] &\leq \sqrt{\var{\frac{1}{N}\sum_{n=1}^N\sqrt{\frac{p(X_n)}{q(X_n)}}}} \\
&= \sqrt{\frac{1}{N}\var{\sqrt{\frac{p(X_n)}{q(X_n)}}}} \\
&= \sqrt{\frac{1}{N}}\sqrt{1 - \left(\int \sqrt{q(x)p(x)}\dee x\right)^2}\\
&= \sqrt{\frac{1}{N}}\sqrt{\hell[2]{p}{q}\left(2-\hell[2]{p}{q}\right)}.
\]
For the second bound, using the triangle inequality, and cancelling out normalization constants
\[
&\EE\left[ \left| \widetilde{\hell[2]{p}{q}}- \hell[2]{p}{q}\right| \right] \\
\leq &
\EE\left[ \frac{\frac{1}{N}\sum_{n=1}^N \sqrt{\frac{p(X_n)}{q(X_n)}}}{\sqrt{\frac{1}{N}\sum_{n=1}^N \frac{p(X_n)}{q(X_n)}}}\left| 1 - \sqrt{\frac{\frac{1}{N}\sum_{n=1}^N \frac{p(X_n)}{q(X_n)}}{\EE\left[\frac{p(X_n)}{q(X_n)}\right]}}\right| \right] + 
\EE\left[ \left| \frac{\frac{1}{N}\sum_{n=1}^N \sqrt{\frac{p(X_n)}{q(X_n)}} - \EE\left[\sqrt{\frac{p(X_n)}{q(X_n)}}\right] }{\sqrt{\EE\left[\frac{p(X_n)}{q(X_n)}\right]}} \right|\right]
\]
By Jensen's inequality on the left term and Cauchy-Schwarz on the right, and noting that $\EE\left[p/q\right] = 1$,
\[
\EE\left[ \left| \widetilde{\hell[2]{p}{q}} - \hell[2]{p}{q}\right| \right] \leq &
\EE\left[ \left| 1 - \sqrt{\frac{1}{N}\sum_{n=1}^N \frac{p(X_n)}{q(X_n)}}\right| \right] + 
\sqrt{\frac{2}{N}}\hell{p}{q}
\]
The left term can be bounded via Cauchy-Schwarz and Jensen's inequality:
\[
\EE\left[ \left| 1 - \sqrt{\frac{1}{N}\sum_{n=1}^N \frac{p(X_n)}{q(X_n)}}\right| \right] &\leq
\sqrt{2 - 2\EE\left[\sqrt{\frac{1}{N}\sum_{n=1}^N \frac{p(X_n)}{q(X_n)}}\right]}\\
&\leq \sqrt{2 - 2\EE\left[\frac{1}{N}\sum_{n=1}^N \sqrt{\frac{p(X_n)}{q(X_n)}}\right]}\\
&= \sqrt{2}\hell{p}{q}
\]
Combining these results yields the second inequality. 
\eprfof

\subsection{Theoretical tools for establishing convergence of \cref{alg:ubvi}}\label{sec:convergenceproofs}
%
%
\bnlem\label{lem:hf}
Define $\hf \defined \argmin_{h\in\closure{\spann{\mcH}} : \|h\|_2 = 1} \|f - h\|_2$. 
Then $\hf$ exists, is unique, and is nonnegative. 
\enlem
\bprfof{\cref{lem:hf}}
Since $\closure{\spann{\mcH}}$ is a closed convex set, there exists a unique function $\hf'$ 
of minimum distance to $f$. Note that $\hf'$ is nonnegative since $f$ is nonnegative,
so otherwise $\hf'$ could be replaced with $\max\{0, \hf'\}$ without increasing the distance to $f$.
Furthermore, the error $\epsilon \defined f - \hf'$ is orthogonal to $\closure{\spann{\mcH}}$.
Since $f$ is not orthogonal to $\closure{\spann{\mcH}}$, $\hf' \neq 0$, so set $\hf = \frac{\hf'}{\|\hf'\|_2}$. Suppose there
is another unit-norm function $g \in \closure{\spann{\mcH}}$ at least as close to $f$; then
\[
0 \geq \left<f, \frac{\hf'}{\|\hf'\|_2} - g\right>
= \left<\hf'+\epsilon, \frac{\hf'}{\|\hf'\|_2} - g\right>
&= \left<\hf', \frac{\hf'}{\|\hf'\|_2} - g\right>\\
&= \|\hf'\|_2 - \left<\hf', g\right>.
\]
Dividing both sides by $\|\hf'\|_2$ yields the inequality
$\left<\hf, g\right> \geq 1$,
implying that $g=\hf$, and thus $\hf$ is unique.
\eprfof
\bnlem\label{lem:tau}
$\tau \leq \frac{\left<\hf, g_1\right>}{1-\sqrt{1 - \left<\hf, g_1\right>^2}} < \infty$.
\enlem
\bprfof{\cref{lem:tau}}
Set $h_1 = \left<\hf, g_1\right>g_1$ where $g_1$ is chosen from \cref{eq:geogreedy},
and $\forall \, i> 1$, set $h_i = 0$. Since $f$ is not orthogonal to $\closure{\spann{\mcH}}$,
$\left<\hf, g_1\right> > 0$, so $\tau < \infty$.
\eprfof
\bnlem\label{lem:descent}
Suppose at each iteration, the optimization in \cref{eq:geogreedy} is solved
with multiplicative error $(1-\delta)$ relative to the optimal objective. 
Then
\[
 J_{n+1} \leq J_n(1-J_n) \text{ where } J_n \defined \left(\frac{1-\delta}{\tau}\right)^2\left(1-\left<\hf, \bg_n\right>^2\right).
\]
\enlem

\bprfof{\cref{lem:descent}}
Taking the derivative of the objective in \cref{eq:protogreedy} with respect to $x$ and setting to zero, the solution is
\[
x^\star &= \sqrt{\frac{\left<f, \frac{h-\left<h, \bg_n\right> \bg_n}{\|h-\left<h, \bg_n\right> \bg_n\|_2}\right>^2}{\left<f, \frac{h-\left<h, \bg_n\right> \bg_n}{\|h-\left<h, \bg_n\right> \bg_n\|_2}\right>^2+\left<f, \bg_n\right>^2}}.\label{eq:geosearchsoln}
\]
Suppose at iteration $n+1$, instead of $g_{n+1}$ we obtain a function $h$ satisfying a $(1-\delta)$-relative approximation to \cref{eq:geogreedy}.
Then using the optimal value for $x^\star$ from \cref{eq:geosearchsoln},
noting that the quadratic weight optimization provides at least as much error reduction as the geodesic update with $x^\star$,
 and noting that $f = \hf' + \epsilon$ where $\epsilon \perp \closure{\spann{\mcH}}$
and $\hf'$ is from the proof of \cref{lem:hf}, we find the recursion
\[
&\|\hf'\|^2_2-\left<\hf', \bg_{n+1}\right>^2 \\
&= \left(\|\hf'\|^2_2-\left<\hf', \bg_n\right>^2\right)\left(1- \left<\frac{\hf' - \left<\hf',\bg_n\right>\bg_n}{\left\|\hf' - \left<\hf',\bg_n\right>\bg_n\right\|}, 
\frac{h - \left<h,\bg_n\right>\bg_n}{\left\|h - \left<h,\bg_n\right>\bg_n\right\|} \right>^2\right)\\
&\leq \left(\|\hf'\|^2_2-\left<\hf', \bg_n\right>^2\right)\left(1- (1-\delta)^2\left<\frac{\hf' - \left<\hf',\bg_n\right>\bg_n}{\left\|\hf' - \left<\hf',\bg_n\right>\bg_n\right\|}, 
\frac{g_{n+1} - \left<g_{n+1},\bg_n\right>\bg_n}{\left\|g_{n+1} - \left<g_{n+1},\bg_n\right>\bg_n\right\|} \right>^2\right).
\]
Now again using the fact that $\epsilon \perp \closure{\spann{\mcH}}$ as well as the fact that $g_{n+1}$ is the 
argmax of \cref{eq:geogreedy}, we can replace $g_{n+1}$ with any convex combination of other elements of $\mcH$, so
\[
&\|\hf'\|^2_2-\left<\hf', \bg_{n+1}\right>^2 \\
&\leq \left(\|\hf'\|^2_2-\left<\hf', \bg_n\right>^2\right)\left(1- (1-\delta)^2\left<\frac{f - \left<f,\bg_n\right>\bg_n}{\left\|\hf' - \left<\hf',\bg_n\right>\bg_n\right\|}, 
\frac{g_{n+1} - \left<g_{n+1},\bg_n\right>\bg_n}{\left\|g_{n+1} - \left<g_{n+1},\bg_n\right>\bg_n\right\|} \right>^2\right)\\
&\leq \left(\|\hf'\|^2_2-\left<\hf', \bg_n\right>^2\right)\left(1- (1-\delta)^2\sup_{h_i\in\cone{\mcH}}\left<\frac{f - \left<f,\bg_n\right>\bg_n}{\left\|\hf' - \left<\hf',\bg_n\right>\bg_n\right\|}, 
\frac{\sum_i h_i - \left<\sum_i h_i,\bg_n\right>\bg_n}{D} \right>^2\right),
\]
where $D = \sum_i \|h_i\| \left\|h_i - \left<h_i,\bg_n\right>\bg_n\right\|$. 
Define $\nu\defined \sum_i h_i - \hf$.  Again using $\epsilon \perp \closure{\spann{\mcH}}$,
and normalizing the left vector by $\|\hf'\|$ yields
\[
&= \left(\|\hf'\|^2_2-\left<\hf', \bg_n\right>^2\right)\left(1- (1-\delta)^2\sup_{h_i\in\cone{\mcH}}\left<\frac{\hf - \left<\hf,\bg_n\right>\bg_n}{\left\|\hf - \left<\hf,\bg_n\right>\bg_n\right\|}, 
\frac{\hf - \left<\hf,\bg_n\right>\bg_n + \nu - \left<\nu,\bg_n\right>\bg_n}{D} \right>^2\right).
\]
Now noting that the inner term is minimized when $\nu = -\|\nu\| \hf$, we have that
\[
&\leq \left(\|\hf'\|^2_2-\left<\hf', \bg_n\right>^2\right)\left(1- \sup_{h_i\in\cone{\mcH}}\frac{(1-\delta)^2(1-\|\nu\|)^2}{D^2}(1-\left<\hf, \bg_n\right>^2)\right).
\]
Finally, dividing both sides by $\|\hf'\|^2_2$ and noting that $D\leq \sum_i \|h_i\|$,
\[
1-\left<\hf, \bg_{n+1}\right>^2 &\leq \left(1-\left<\hf, \bg_n\right>^2\right)\left(1- \left(\frac{1-\delta}{\tau}\right)^2\left(1-\left<\hf, \bg_n\right>^2\right)\right).
\]
Denoting $J_n = \left(\frac{1-\delta}{\tau}\right)^2\left(1-\left<\hf, \bg_n\right>^2\right)$,
and multiplying both sides by $\left(\frac{1-\delta}{\tau}\right)^2$ yields the recursion
\[
J_{n+1} &\leq J_n\left(1- J_n\right).
\]
\eprfof
\bprfof{\cref{thm:convergence}}
By \citep[Lemma A.6]{Campbell19},
the recursion \cref{lem:descent} satisfies
$J_n \leq \frac{J_0}{1+J_0n}$. 
Substituting the definition of $J_n$ and noting that 
$\hell{\hp}{q_n}^2 = 1-\left<\hf, \bg_n\right> \leq 1-\left<\hf, \bg_n\right>^2$
yields the result. 
\eprfof

\end{document}